\documentclass[11pt]{article}
\pdfoutput=1
\usepackage{jcappub}
\usepackage{slashed}
\usepackage{amsmath}
\usepackage[frak=euler,scr=boondox,bb= pazo]{mathalfa}
\DeclareMathOperator{\sech}{sech}

\def\be{\begin{equation}}
\def\ee{\end{equation}}
\def\bea{\begin{eqnarray}}
\def\eea{\end{eqnarray}}

\begin{document}

\title{Towards a theory of machine learning} 

\author{Vitaly Vanchurin}

\emailAdd{vvanchur@d.umn.edu}

\date{\today}

\affiliation{Department of Physics, University of Minnesota, Duluth, Minnesota, 55812 \\
Duluth Institute for Advanced Study, Duluth, Minnesota, 55804}

\abstract{We define a neural network as a septuple consisting of (1) a state vector, (2) an input projection, (3) an output projection, (4) a weight matrix,  (5) a bias vector, (6) an activation map and (7) a loss function. We argue that the loss function can be imposed either on the  boundary (i.e. input and/or output neurons) or in the bulk  (i.e. hidden neurons) for both supervised and unsupervised systems. We apply the principle of maximum entropy to derive a canonical ensemble of the state vectors subject to a constraint imposed on the bulk loss function by a Lagrange multiplier (or an inverse temperature parameter). We show that in an equilibrium the canonical partition function must be a product of two factors: a function of the temperature and a function of the bias vector and weight matrix. Consequently, the total Shannon entropy consists of two terms which represent respectively a thermodynamic entropy and a complexity of the neural network. We derive the first and second laws of learning: during learning the total entropy must decrease until the system reaches an equilibrium (i.e. the second law), and the increment in the loss function must be proportional to the increment in the thermodynamic entropy plus the increment in the complexity (i.e. the first law).  We calculate the entropy destruction to show that the efficiency of learning is given by the Laplacian of the total free energy which is to be maximized in an optimal neural architecture, and explain why the optimization condition is better satisfied in a deep network with a large number of hidden layers. The key properties of the model are verified numerically by training a supervised feedforward neural network using the method of stochastic gradient descent. We also discuss a possibility that the entire universe on its most fundamental level is a neural network. }

\maketitle

\section{Introduction}

Despite of many attempts \cite{Saxe,Choromanska, Kadmon,Advani, LinTegmarkRolnick, Bottleneck2} the effectiveness of deep learning has so far no clear explanation. This is rather surprising given that a neural network is a very simple and a well-defined mathematical object \cite{Galushkin, Schmidhuber, Haykin}. What makes it difficult to analyze is that the deep neural networks are typically described with a very large number of parameters, e.g. weight matrix, bias vector, training data, etc. For such systems most of the analytical techniques are not very useful and one must rely on numerics. The situation is very similar to what  happens in physics. Physical systems (both classical and quantum) can often be solved exactly when the number of degrees of freedom is small, but the problem becomes intractable when the number of degrees of freedom is large. Fortunately, there is a set of ideas which proved very useful for analyzing physical systems with many degrees of freedom. It is statistical mechanics. The main point of the present paper is to apply the methods of statistical mechanics to machine learning. Note that the idea of using statistical methods to study artificial neural networks is not new and goes back to the works of Hopfield \cite{Hopfield, Hopfield2} (see Ref.  \cite{Carleo:2019ptp} for a recent review of statistical methods used in machine learning as well as applications of machine learning methods in physics), but in this paper we shall take one step further and develop both equilibrium and non-equilibrium thermodynamic descriptions of neural networks. In the remainder of this section, we will summarize the main results as it might help the reader to navigate through the paper.

In Sec. \ref{sec:septuple} we set the stage by defining a neural septuple (i.e. state vector, input/output projection operators, weight matrix,  bias vector,  activation map and loss function). The septuple is not equivalent to the standard neural architecture used in machine learning, but it does include such systems as a special limit. There are three main motivations to define these more general structures. First of all, we want to develop a unified treatment of different types of learning algorithms, i.e. supervised, unsupervised, etc. Secondly, we want the very structure of hidden layers to be a dynamical variable in addition to the weight matrices and bias vectors. And finally, we want to have a theoretical framework which is suitable for a statistical description. 

In Sec. \ref{sec:unsupervised} we address the main problem of unsupervised learning, namely, what should be an appropriate loss function if the training dataset contains only input, but no output data. We claim that an answer can be obtained by defining a local error and a local objective for hidden neurons (or in the bulk) instead of a more conventional error for output neurons (or on the boundary). The boundary loss is usually given by a sum over errors on the boundary (i.e. over input/output neurons), but the bulk loss could be a sum over both local errors and local objectives in the bulk (i.e. over hidden neurons). A simple example of a local objective for a neuron is a binary classification of an incoming signal, and then an outgoing signal with values closer to lower- and upper-bounds are rewarded and values in-between are penalized.

In Sec. \ref{sec:ensemble} we consider two statistical ensembles over state vectors: a micro-canonical-type ensemble and a canonical-type ensemble. We expect that in the limit of a large number of neurons the two ensembles are equivalent, as is usually the case in statistical physics, but the latter ensemble (i.e. canonical ensemble) is a lot easier to handle analytically. Moreover, we show that the canonical ensemble can be derived from the Jaynes' maximum entropy principle \cite{Jaynes,Jaynes2}. The principle states that the probability distribution (in our case statistical ensemble) which best represents the current state of knowledge is the one with the largest Shannon entropy. 

In Sec. \ref{sec:partition} we define perhaps the most important object in statistical mechanics - a partition function. For a bulk loss with (or without) a quadratic local objective the canonical ensemble can be approximated by a Gaussian integral and then the (canonical) partition function can be calculated analytically.  A minor complication is that the range of integration (which is set by the range of an activation function) is finite in contrast to the Gaussian integral whose range is infinite. Nevertheless, the problem can be solved by replacing the sharp cut-off on the boundaries of integration with a smooth Gaussian window function. In this section we also define an operator $\hat{G}$ whose spectrum determines the canonical partition function and plays a central role in everything that follows.

In Sec. \ref{sec:equilibrium} we define a time-invariant state of equilibrium (or what we call a learning equilibrium) and show that in such state the partition function must factorize into product of two factors: a function of the temperature and a function of the bias vector and weight matrix.  Among other things it implies that the total free energy is a difference of two terms: a familiar thermodynamic free energy and an unfamiliar product of the temperature and a complexity function. While the thermodynamic free energy is a function of only temperature and usually decreases, the total free energy is expected to increase with learning due to a decrease in the complexity function. This might sound odd, but, in fact, is a mare consequence of an openness of the learning system where the entropy is flowing out of the system.

In Sec. \ref{sec:thermodynamics} we calculate the total entropy of the canonical ensemble and argue that in an equilibrium it must be a sum of a familiar thermodynamic entropy and of an unfamiliar complexity function which is directly related to a dynamical dimensional reduction of the state space. We then argue that the total entropy must decrease until the systems reaches an equilibrium (i.e. the second law of learning \eqref{eq:second_law}) and the increment in the loss function must be proportional to the increment in the thermodynamic entropy plus the increment in the complexity (i.e. the first law of learning  \eqref{eq:first_law}). 

In Sec. \ref{sec:architecture} we calculate a non-equilibrium production of the Shannon entropy (of a probability distribution function over weight matrices and bias vectors) and argue that in an optimal neural architecture the entropy production must be maximized (or, more precisely, the entropy destruction should be minimized). This is in a complete agreement with the so-called stationary entropy production principle that was used in \cite{entropic} to derive an approximate Schr\"odinger equation from a highly constrained  stochastic process and in \cite{emergentgravity} to derive an approximate Einstein equation from non-equilibrium thermodynamics of the metric tensor. We then show that the rate of the entropy production is proportional to the Laplacian of the free energy in the configuration space of the weight matrices and bias vectors.

In Sec. \ref{sec:deep_shallow} we used the criteria of minimizing the entropy destruction (or minimizing the negative Laplacian of the free energy) to derive an expected dynamics of a spectrum of operator $\hat{G}$ in an optimal neural architecture. In particular, we show that most of the eigenvalues of the operator $\log \hat{G}$ should remain near zero, a small fraction of the largest eigenvalues should move to positive values $\gtrsim 0$ and a very small fraction of eigenvalues should move to very small values $\ll 0$. This implies that the effectiveness of a neural network can be translate into a skewness of the distribution of  eigenvalues of  $\log \hat{G}$, i.e. the larger the skewness the better a neural network is expected to perform. Then we show that the skewness in a deep architecture is much larger than in a shallow architecture which demonstrates why the deep architecture is preferred. 

In Sec. \ref{sec:numerics} we discuss the main results of numerical experiments conducted using the TensorFlow Python library  \cite{TensorFlow} and MNIST database of handwritten images \cite{MNIST}. Two different neural networks (a deep network with two hidden layers and a shallow network with a single hidden layer) were trained using the method of stochastic gradient descent and then the numerical data were analyzed in context of the analytical calculations carried out in the paper.  More specifically, the training evolution of the bulk and boundary loss functions progressed as expected, the predicted dynamics of the spectrum of operator  $ \hat{G}$ was established, and the anticipated relaxation of the complexity function towards equilibrium was  confirmed.

And finally in Sec. \ref{sec:entropic} and Sec. \ref{sec:gravity} we discuss a possibility that the entire universe on its most fundamental level is a neural network.  For such an ambitious proposal to actually work we claim that the three components: quantum mechanics, general relativity and macroscopic observers must all emerge from a microscopic neural network. For the time being, we leave aside the problem of observers (see, however, \cite{Tononi}) and study a possible emergence of quantum mechanics and general relativity. In particular, we show that approximate Schr\"odinger equation  (see Sec. \ref{sec:entropic})  and Einstein equations  (see Sec. \ref{sec:gravity}) can indeed emerge from a network with a large number of neurons not too far from a learning equilibrium. 

\section{Neural septuple}\label{sec:septuple}

We start by introducing all of the essential ingredients of a neural network or, what we shall call, a neural septuple consisting of:

(1) ${\bf x}$, a state vector which describes the state of all neurons, 

(2) $\hat{P}_{in}$, an input projection which describes a subspace spanned by input neurons,

(3) $\hat{P}_{out}$, an output projection which describes a subspace spanned by output neurons,  

(4) $\hat{w}$,  a weight matrix  which describes directed connections between all pairs of neurons,

(5) ${\bf b}$, a bias vector which describes biases in the  inputs of all neurons, 

(6) ${\bf f}$ , an activation map which describes a non-linear transformation, and 

(7) ${H}$, a loss function which describes a learning objective of the entire network. \\
Consider a collection of $N$ neurons described by a column\footnote{We adopt the physicists' notations where the state vector is a column vector and not a row vector which is usually used in the machine learning literature.} {\it state vector}, ${\bf x} \in \mathbb{R}^N $, whose components are real numbers, $x_i \in \mathbb{R} $, but one can also generalize the construction to complex numbers or other fields. There are three types of neurons: input neurons, hidden neurons and output neurons, and so it is convenient to define three subspaces of the state space: input subspace ${\cal V}_{in}$, output subspace ${\cal V}_{out}$ and hidden subspace  ${\cal V}_{hid}$. We shall also refer to the direct sum of the input and output subspaces, ${\cal V}_{in}\oplus {\cal V}_{out}$,  as a boundary and to the hidden subspace, ${\cal V}_{hid}$, as a bulk. A neural network is trained by specifying the components of ${\bf x}$ in the boundary subspace which represent only  the input and output neurons. The boundary components can be described with two projection operators (or matrices): an {\it input projection} $\hat{P}_{in}$ and an {\it output projection} $\hat{P}_{out}$. These operators can be used to project a state vector ${\bf x}$ to either input subspace, i.e. $\hat{P}_{in}{\bf x} \in {\cal V}_{in}$, output subspace, i.e. $\hat{P}_{out}{\bf x} \in {\cal V}_{out}$, boundary subspace, i.e. $( \hat{P}_{in}+\hat{P}_{out}  ) {\bf x} \in {\cal V}_{in}\oplus {\cal V}_{out}$, or bulk subspace, i.e. $(\hat{I} - \hat{P}_{in}-\hat{P}_{out}  ) {\bf x} \in {\cal V}_{hid}$.

The neurons are connected into a neural network with connections described by a {\it weight matrix}, $\hat{w}$, which is also an adjacency matrix of a weighted directed graph with individual neurons representing the nodes of the graph. For the neural networks considered here the components of the weight matrix are assumed to be real numbers, $w_{ij}  \in \mathbb{R}$, but one can also generalize the construction to complex numbers or other fields. In addition, for the so-called feedforward neural network with $L$ layers (i.e. one input layer, one output layer and $L-2$ hidden layers) the weight matrix is taken to be nilpotent, i.e.
 \be
 \hat{w}^n = \left ( \hat{w}^T \right )^n = 0 \;\;\;\forall n \ge L, \label{eq:nilpotent}
 \ee
there are no incoming connections to the input neurons, i.e. 
\be
\hat{P}_{in} \hat{w}=0,
\ee 
and there are no outgoing connections from the output neurons, i.e. 
\be
\hat{P}_{out} \hat{w}^T=0.
\ee 
The state vector can only change when either a new training data $\hat{P}_{in} {\bf x}_\partial \in {\cal V}_{in}$ are entered, 
\bea
{\bf x}(0) =\hat{P}_{in} {\bf x}_\partial \label{eq:initial_conditions}
\eea
or the new data propagate through the network
\be
{\bf x}({t+1}) = {\bf f} \left ( \hat{w} {\bf x}(t)+ {\bf b} \right) \label{eq:eom}.
\ee
The vector ${\bf b} \in \mathbb{R}^N$ is a {\it bias vector} and  ${\bf f}: \mathbb{R}^N  \rightarrow \mathbb{R}^N$ is an {\it activation map} which acts separately on each component, i.e.
\be
f_i( {\bf y}) = f_i\left (y_i \right ). \label{eq: activation}
\ee
where $f_i(y)$'s are the activation functions  (e.g. $f_i(y) = \tanh(y)$, $f_i(y) =\max(0,y)$).

For the input neurons with no incoming connections to remain fixed, i.e. 
\be
\hat{P}_{in} {\bf x}({t}) = \hat{P}_{in} {\bf x}({0})=  {\bf x}({0})=\hat{P}_{in} {\bf x}_\partial, \label{eq:input_constraint} 
\ee
an additional condition must also be imposed on the input bias, 
\be
{\bf f} \left (\hat{P}_{in}  {\bf b}  \right )= \hat{P}_{in} {\bf x}_\partial .\label{eq:bias_constraint}
\ee
This condition is satisfied for a feedforward neural network, but need not be satisfied for more general learning systems. After a finite  number of steps $t$ the state vector ${\bf x}(t)$ may converge to a fixed state ${\bf x}(t)=\bar{\bf x}$ defined by a fixed point equation
\be
\bar{\bf x} = {\bf f} \left ( \hat{w} \bar{\bf x}+ {\bf b} \right). \label{eq:fixed_point}
\ee
For example, in a deep feedforward neural network with $L$ layers the fixed state would be reached after $L-1$ steps, i.e. ${\bf x}({L-1})=\bar{\bf x}$, given that the condition on the input bias \eqref{eq:bias_constraint} is satisfied. For more general systems the state may or may not converge to a fixed point depending on the  activation transformation \eqref{eq:eom} and initial conditions \eqref{eq:initial_conditions}. 

The final ingredient of a neural  septuple is a loss function. In a feedforward neural network the loss function is usually defined by projecting the fixed state $\bar{\bf x}$ to the output subspace $\hat{P}_{out} \bar{\bf x} \in {\cal V}_{out}$ and then by comparing the result with a desired output state  $\hat{P}_{out} {\bf x}_\partial  \in {\cal V}_{out}$. For example, one can define a loss function as a squared error of the output neurons,    
\bea
H_{\partial}(\bar{\bf x}, {\bf b}, \hat{w}) & = & \left ( \hat{P}_{out} \bar{\bf x} - \hat{P}_{out} {\bf x}_\partial \right )^T \left ( \hat{P}_{out} \bar{\bf x} - \hat{P}_{out} {\bf x}_\partial \right ) \label{eq:square_error}\\
& = & \left ( \bar{\bf x} - {\bf x}_\partial \right )^T  \hat{P}_{out}^T  \hat{P}_{out} \left ( \bar{\bf x} - {\bf x}_\partial \right )\notag\\
& = & \left ( \bar{\bf x} - {\bf x}_\partial \right )^T   \hat{P}_{out} \left ( \bar{\bf x} - {\bf x}_\partial \right )\notag
\eea
where in the last line we used that the projection operator is orthogonal, i.e. 
\be
\hat{P}_{out}^T  \hat{P}_{out} = \hat{P}^2_{out} =\hat{P}_{out}.
\ee 
Since there is no error on the input neurons \eqref{eq:input_constraint} we can also rewrite it as a squared error on all boundary (i.e. input and output) neurons
\be
H_\partial(\bar{\bf x},{\bf b}, \hat{w}) =  \frac{1}{2} \left ( \bar{\bf x} -  {\bf x}_\partial \right )^T   (  \hat{P}_{in}+ \hat{P}_{out} ) \left (  \bar{\bf x} - {\bf x}_\partial \right ). \label{eq:boundary_error}
\ee
For this reason, we shall refer to $H_\partial$ as a boundary loss function.

\section{Supervised vs. unsupervised}\label{sec:unsupervised}

In the pervious section we defined a neural network as a neural septuple $\left ( {\bf x}, \hat{P}_{in},  \hat{P}_{out},  \hat{w} , {\bf b}, {\bf f}, H \right )$ where $ {\bf x}$ is a state vector of all (input, output and hidden) neurons, $ \hat{P}_{in}{\bf x}$ is a state of only input neurons, $ \hat{P}_{out}{\bf x}$ is a state of only output neurons, $ \hat{w}$ is a weight matrix between all pairs of neurons, ${\bf b}$ is a bias vector for all neurons, ${\bf f}({\bf y})$ is an activation map and $H({\bf x}, {\bf b}, \hat{w})$ is a loss function. A simple example of a loss function is the boundary loss \eqref{eq:boundary_error} which is known to work very well in a supervised learning. Unfortunately, the boundary loss cannot be used in unsupervised systems where the output subspace is empty, ${\cal V}_{out}= \emptyset$, and thus the boundary loss is always zero, $H=H_\partial=0$.\footnote{In our description an auto-encoder is viewed as a supervised system with periodic boundary conditions, i.e. the input and output states are set equal to each other.} For this reason, in unsupervised systems (beyond auto-encoders) we must consider other loss functions which are, perhaps, more general than the boundary loss.

A key observation is that in equation \eqref{eq:boundary_error} the boundary loss was due to a mismatch in the output conditions or (together with input conditions) in  the boundary conditions, i.e. $(  \hat{P}_{in}+ \hat{P}_{out})\bar{\bf x} \neq {\bf x}_\partial$, but the fixed point equation \eqref{eq:fixed_point} was satisfied exactly. Alternatively, we can assume that the boundary conditions are satisfied exactly  $ (  \hat{P}_{in}+ \hat{P}_{out} ) {\bf x} = {\bf x}_\partial$, but the fixed point equation is only approximate. Then we can define a bulk loss (as opposed to the boundary loss) as a sum of squares of errors in the fixed point equation, i.e.
\be
H(\bar{\bf x}, {\bf b}, \hat{w}) =   \frac{1}{2} \left (  \bar{\bf x}  - {\bf f} \left ( \hat{w} \bar{\bf x}+ {\bf b} \right) \right )^T \left (  \bar{\bf x}  - {\bf f} \left ( \hat{w} \bar{\bf x}+ {\bf b} \right) \right ), \label{eq:bulk_error}
\ee 
where $\bar{\bf x}$ is the value of $\bf x$ at a minimum of  $H({\bf x}, {\bf b}, \hat{w})$ subject to boundary conditions $ (  \hat{P}_{in}+ \hat{P}_{out} ) {\bf x} = {\bf x}_\partial$, i.e.
\be
H(\bar{\bf x}, {\bf b}, \hat{w}) = \min_{ (  \hat{P}_{in}+ \hat{P}_{out} ) {\bf x} = {\bf x}_\partial} H({\bf x}, {\bf b}, \hat{w}). \label{eq:minima}
\ee
This is the simplest bulk loss\footnote{This bulk loss function is similar in spirit (but not the same) to the error calculated in the back-propagation  algorithm where the error on the output neurons is back-propagated to the hidden neurons.}  which is still zero for unsupervised feedforward neural networks, but can be easily generalized to functions which can be used in both supervised and unsupervised learning. 

The main idea is that, from the point of view of an individual neuron, a (more general) learning objective can be modeled as a minimization of a local error and at the same time a maximization of a local objective. It is convenient to think of the local error as a supervised quantity (e.g. $(\bar{x}_i - f_i ( \sum_{j} w_{ij} \bar{x}_j + b_i ))^2$ in the bulk loss function \eqref{eq:bulk_error}) and of the local objective as a (yet to be defined) unsupervised quantity.  Then even if the local error is already at its minimum (as is always the case for unsupervised feedforward neural networks) there is still another quantity which needs to be extremized, i.e. the local objective. This does not mean that the inclusion of the local objective would only benefit an unsupervised learning. Once an appropriate local objective is identified it can be incorporated into a (bulk or boundary)  loss function to improve the convergence of a learning algorithm.  

For example, the local objective might be a binary classification of an incoming signal $\sum_{j} w_{ij} \bar{x}_j + b_i$ and then the values of $\bar{x}_i$ closer to lower- and upper-bounds should be rewarded and values in-between should be penalized. Such a classification objective can always be modeled with an appropriately chosen ``potential'' term for each neuron. For example if
\be
{ V}(\bar{\bf x}) = \sum_i V_i(\bar{x}_i) = - \sum_i \frac{m}{2}  \bar{x}_i^2 \label{eq:tachyon}
\ee
then the bulk loss function can be defined as
\bea
H(\bar{\bf x}, {\bf b}, \hat{w}) &=&  \frac{1}{2} \left (  \bar{\bf x}  - {\bf f} \left ( \hat{w} \bar{\bf x}+ {\bf b} \right) \right )^T \left (  \bar{\bf x}  - {\bf f} \left ( \hat{w} \bar{\bf x}+ {\bf b} \right) \right ) +  { V}(\bar{\bf x}) \label{eq:bulk_loss} \\
&=& \frac{1}{2}  \sum_i  \left[ \left (\bar{x}_i - f_i \left ( \sum_{j} w_{ij} \bar{x}_j + p_j \right ) \right )^2  - {m}\bar{x}_i^2  \right ].\notag
\eea
By minimizing this loss function we accomplish both tasks: the minimization of the local error and maximization of the local objective (in this case the binary classification objective). Note that the ``tachyonic  potential'' \eqref{eq:tachyon} does not lead to any runaway solutions if the range of $\bar{x}_i$ is bounded by the range of the activation function. For example, if the activation function is $f(y)=\tanh(y)$ then $x_i \in (-1,1)$.  More generally, any two (or more) neurons might have a common objective and then the potential term must also include ``interactions'', i.e. ${V} (\bar{\bf x}) =\sum_i V_i(\bar{x}_i)+ \sum_{ij} g^{ij} \bar{x}_i \bar{x}_j ...\;$. In either case, according to \eqref{eq:minima}, the corresponding learning objective remains the same, i.e. we must adjust $\hat{w}$ and ${\bf b}$ in such a way that for a given set of boundary condition  $(  \hat{P}_{in}+ \hat{P}_{out} )  \bar{\bf x}={\bf x}_\partial$ the bulk loss function is minimized. What is, however, different is that the bulk loss function  $H({\bf x}, {\bf b}, \hat{w})$ is now given by \eqref{eq:bulk_loss} which contains both a local error (a supervised or a kinetic term) and a local objective (an unsupervised or a potential term). As a result, the corresponding bulk loss function is well-defined and (generically) non-zero for both supervised and unsupervised systems.  Unfortunately, the exact solutions of equation \eqref{eq:minima} are difficult to obtain and so the statistical methods must be employed.

\section{Statistical ensembles}\label{sec:ensemble}

There are basically two ways to proceed: experimental (based on numerics) or theoretical (based on statistics). We will start with statistical approach as it might assist us in numerical searches. For starters, consider a statistical ensemble of boundary conditions or, more precisely, a probability distribution ${p}_\partial({\bf x}_\partial)$ over components of the state vector in the boundary subspace ${\bf x}_\partial=(  \hat{P}_{in}+ \hat{P}_{out} )  {\bf x}$. Such a distribution can, for example, be extracted from a training dataset. Then, instead of minimizing a loss function for individual boundary data, the learning objective could be to minimize an ensemble-averaged loss function, i.e.
\be
 U_0({\bf b}, \hat{w}) \equiv  \int d^{N_\partial} x_\partial  \min_{(  \hat{P}_{in}+ \hat{P}_{out} )  {\bf x} = {\bf x}_\partial} H({\bf x}, {\bf b}, \hat{w}){p}_\partial({\bf x}_\partial) 
\ee
where $N_\partial \le N$ is the dimensionality of the boundary subspace. If we now extend the probability distribution into the bulk by defining 
\be
{p}_0({\bf x}) = {p}_\partial((  \hat{P}_{in}+ \hat{P}_{out} )  {\bf x}) \delta\left (\bar{\bf x}-{\bf x}\right ),\label{eq:microcanonical}
\ee
where $\bar{\bf x}$ is given by \eqref{eq:minima}, then the ensemble-averaged loss function is simply
\be
 U_0({\bf b}, \hat{w})  = \int d^{N} x  H({\bf x}, {\bf b}, \hat{w}){p}_0({\bf x}). \label{eq:constraint0}
\ee 
Of course, all we did was to move the difficulty of calculating $\bar{\bf x}$ into ${p}_0({\bf x})$, but that does not solve the main problem. It is still a computationally intensive task to calculate $ U_0({\bf b}, \hat{w})$ exactly and this is where statistical mechanics comes to rescue. The key idea is to replace the micro-canonical-type ensemble  \eqref{eq:microcanonical} with a canonical-type ensemble \eqref{eq:canonical}. Note that for sufficiently large systems (in our case large neural networks) one can often show that the two ensembles are equivalent (i.e. predictions are identical) but the canonical ensemble is much easier to handle analytically.

Consider a statistical ensemble of neural networks, or a probability distribution ${p}({\bf x})$, over state vectors ${\bf x}$. Let's say we do not know how the network was trained (i.e. which algorithm was used), but we do know that the ensemble-averaged bulk loss $H({\bf x}, {\bf b}, \hat{w})$ was reduced to some fixed value,
\be
 U({\bf b}, \hat{w})  = \int d^{N} x  H({\bf x}, {\bf b}, \hat{w}){p}({\bf x}). \label{eq:constraint1}
\ee
Then according to the principle of maximum entropy \cite{Jaynes, Jaynes2}, the most reasonable distribution ${p}({\bf x})$ is a distribution  which has the largest Shannon entropy 
\be
S({p}) \equiv - \int d^{N} x \;  {p}({\bf x}) \log {p}({\bf x}) =  - \langle \log {p}({\bf x}) \rangle . \label{eq:entropy}
\ee 
subject to constraint \eqref{eq:constraint1}. The maximization problem can be solved using the method of Lagrange multipliers. If we define a ``Lagrangian'' 
\bea
{L}({p}, \beta, \nu)  &=&  S({p}) + \beta \left ( U({\bf b}, \hat{w})  -  \int d^{N} x \; {p}({\bf x}) H ({\bf x}, {\bf b}, \hat{w}) \right ) +\nu \left (1- \int d^{N} x  \; {p}({\bf x})  \right ),\notag\\
&=& \int d^{N} x  \; {p}({\bf x})  \left ( - \log  {p}({\bf x})  - \beta H({\bf x}, {\bf b}, \hat{w})    - \nu \right ) + \beta  U({\bf b}, \hat{w})  +\nu\label{eq:Lagrangian}
\eea
then at a maximum of $ L({p}, \beta, \nu)$ the variations with respect to ${p}({\bf x}) $, $\beta$ and $\nu$ must vanish
\bea
0 &=& \frac{\delta {L}({p}, \beta, \nu)}{\delta {p}({\bf x}) } =   - \beta H({\bf x}, {\bf b}, \hat{w}) - \log {p}({\bf x})  - 1 - \nu\\
0 &=&  \frac{\partial { L}({p}, \beta, \nu)}{ \partial \beta} =  U({\bf b}, \hat{w}) - \int d^{N} x \;{p}({\bf x}) H({\bf x}, {\bf b}, \hat{w}) \label{eq:constraint2}\\
0 &=&  \frac{\partial { L}({p}, \beta, \nu)}{ \partial \nu} = 1 - \int d^{N} x \; {p}({\bf x}). \label{eq:constraint3}
\eea
Therefore, the maximum entropy distribution must be given by 
\be
{p}({\bf x})  = \exp\left (- \beta H({\bf x}, {\bf b}, \hat{w})-1-\nu \right ) \label{eq:canonical}
\ee
with  Lagrange multipliers  $\beta$  and $\nu$ determined from the constraint \eqref{eq:constraint2} and  normalization condition \eqref{eq:constraint3}. In what follows we shall refer to the distribution \eqref{eq:canonical} as a canonical ensemble.

\section{Partition function}\label{sec:partition}

The partition function for the canonical ensemble \eqref{eq:canonical} is defined as ${\cal Z}\equiv \exp(1+\nu)$ and can be expressed as an integral over the state space
\bea
{\cal Z}(\beta, {\bf b}, \hat{w}, ...) &=& \int d^{N} x  \;e^{- \beta H\left  ( {\bf x}, {\bf b}, \hat{w} \right )} \label{eq:canonical_partition}
\eea
where $...$ should remind us of any additional variables (e.g. $m$) which could determine the functional form of $ H$.  To calculate the partition function \eqref{eq:canonical_partition} for a bulk loss \eqref{eq:bulk_loss} we can approximate the integral with a Gaussian. This can be done by first expanding the activation function \be
{\bf f}\left ( \hat{w} {\bf x}+ {\bf b}\right ) = {\bf f} (\hat{w} \langle {\bf x} \rangle + {\bf b} ) + \hat{f}' \hat{w} ({\bf x} - \langle {\bf x}\rangle) + {\cal O}(({\bf x} - \langle {\bf x}\rangle)^2)\label{eq:expansion}
\ee
where the ensemble-averaged state vector is
\be
\langle {\bf x}\rangle  \equiv \int d^{N} x \; {\bf x}\; {p}({\bf x})
\ee
and a diagonal matrix of first derivatives of the activation function is
\be
f'_{ii} \equiv \left (\frac{d f(y_i)}{d y_i}\right )_{y_i = \sum_j w_{ij} \langle x_j \rangle + b_i}.  \label{eq:F}
\ee
Then to the first order in perturbation theory the bulk loss function is
\bea
H({\bf x}, {\bf b}, \hat{w}) &\approx&    \frac{1}{2} ({\bf x} - \langle {\bf x}\rangle) ^T\hat{G} ({\bf x} - \langle {\bf x}\rangle) - \frac{m}{2}{\bf x}^T  {\bf x}  \label{eq:approximation}
\eea
where
\be
\hat{G} \equiv \left ( \hat{I} - \hat{f}' \hat{w}\right)^T \left ( \hat{I} - \hat{f}' \hat{w}\right).
\ee
Next, we note that the domain of ${\bf x}$ is bounded by the range of the activation function. For example, if the activation function is $f(x)=\tanh(x) \in (-1,1)$, then the partition function is
\be
{\cal Z}(\beta, {\bf b}, \hat{w}) =   \int_{{\bf x} \in (-1,1)^N} d^{N} x\, e^{- \beta H({\bf x}, {\bf b}, \hat{w})}.
\ee
The sharp boundaries can be approximated with a smooth Gaussian window function, i.e.
\be
{\cal Z}(\beta, {\bf b}, \hat{w}) \approx    \int_{{\bf x} \in (-\infty ,\infty)^N} d^{N} x\, e^{- \beta H({\bf x}, {\bf b}, \hat{w})} e^{ - \frac{1}{2}{\bf x}^T  {\bf x}  }.
\ee
and then the overall partition function is a Gaussian and can be easily evaluated, 
\bea
{\cal Z}(\beta, {\bf b}, \hat{w}) &\approx &   \int d^{N} x\, e^{- \beta H({\bf x}, {\bf b}, \hat{w}) - \frac{1}{2}{\bf x}^T  {\bf x} } \approx   \int d^{N} x\, e^{- \frac{\beta }{2} ({\bf x} - \langle {\bf x}\rangle) ^T\hat{G} ({\bf x} - \langle {\bf x}\rangle) - \frac{1- \beta m}{2}{\bf x}^T  {\bf x}  } \label{eq:R2} \\
&\approx&  (2 \pi)^{N/2} \det \left ( \hat{I} (1-\beta m) +  \beta\hat{G} \right )^{-1/2} \,\exp \left ( { -\frac{1}{2}\langle {\bf x}\rangle^T \left (\frac{(1-\beta m) \beta\hat{G} }{ \hat{I} (1-\beta m) + \beta \hat{G}} \right )  \langle {\bf x}\rangle}\right ).  \notag
\eea
The spectrum of $\hat{G}$ is defined by an eigenvalue equation
\be
\hat{G} {\bf v}_i = \lambda_i {\bf v}_i 
\ee
 where $\lambda_i$ are the real eigenvalues and ${\bf v}_i$ are the respective eigenvectors.  Then the log of partition function is
 \bea
\log {\cal Z}(\beta, {\bf b}, \hat{w})  &\approx& -\frac{1}{2} \sum_i \log \left (  1-\beta m +  \beta \lambda_i \right )  -\frac{1}{2} \sum_i  \frac{(1-\beta m) \beta \lambda_i a_i^2}{1-\beta m+\beta \lambda_i} + \frac{N}{2} \log (2\pi) \;\;\;\;\; \label{eq:logZZ}
\eea
where
\be
\langle {\bf x} \rangle = a_i {\bf v}_i.
\ee
In the limit of a large number of neurons $N$, the average components are small $a_i^2 \ll 1$, the second term in \eqref{eq:logZZ} is subdominant in comparison to the first term and can be dropped. Then the partition function is given by  \eqref{eq:R2}  but without an exponential factor, i.e.
\be
{\cal Z}(\beta, {\bf b}, \hat{w}) \approx  (2 \pi)^{N/2} \det \left ( \hat{I} (1-\beta m) +  \beta\hat{G} \right )^{-1/2} \label{eq:Z}
\ee
 and the log of the partition function is
 \bea
\log {\cal Z}(\beta, {\bf b}, \hat{w})  &\approx& -\frac{1}{2} \log \det \left ( \hat{I} (1-\beta m) +  \beta\hat{G} \right )  + \frac{N}{2} \log (2\pi)\notag\\
&\approx& -\frac{1}{2} Tr \log \left ( \hat{I} (1-\beta m) +  \beta\hat{G} \right )  + \frac{N}{2} \log (2\pi)\label{eq:logZ}\\ \notag
  &\approx& -\frac{1}{2} \sum_i \log \left ( 1 -\beta m +  \beta \lambda_i \right )  + \frac{N}{2} \log (2\pi).
\eea
Note, however, that this is a very rough estimate of the true partition function borne out of our statistical description, but the hope is that this approximation is rich enough to explain at least some aspects of machine learning. 

\section{Learning equilibrium}\label{sec:equilibrium}

Given the partition function \eqref{eq:canonical_partition} the average bulk loss can be calculated by simple differentiation,  
\be
 U(\beta , {\bf b}, \hat{w})  = \int d^{N} x  H({\bf x}, {\bf b}, \hat{w}){p}({\bf x}) = - \frac{\partial}{\partial \beta}  \log({\cal Z}(\beta, {\bf b}, \hat{w})), 
\ee 
where we have explicitly shown that  $U(\beta , {\bf b}, \hat{w})$ depends on the Lagrange multiplier $\beta$ (or, what we shall call, an inverse temperature parameter).  If the neural network was already trained for a very long time, then the weight matrix and the bias vector must be in a state which minimizes the average loss $U(\beta , {\bf b}, \hat{w})$  and then its variations with respect to $\hat{w}$ and ${\bf b}$ must vanish, 
 \bea
 \frac{\partial U(\beta , {\bf b}, \hat{w})  }{\partial w_{ij}} &=& \frac{\partial^2}{\partial w_{ij} \partial \beta}  \log({\cal Z}(\beta, {\bf b}, \hat{w})) =0 \notag\\
 \frac{\partial U(\beta , {\bf b}, \hat{w})  }{\partial b_{i}} &=& \frac{\partial^2}{\partial b_{i} \partial \beta}  \log({\cal Z}(\beta, {\bf b}, \hat{w})) =0. \label{eq:equilibrium}
 \eea
 We shall call this state, a state of the learning equilibrium or just the equilibrium state. 
 
A very important property of the equilibrium, which follows from \eqref{eq:equilibrium}, is that the partition function must be a product of two terms
 \be
{\cal Z}(\beta, {\bf b}, \hat{w}) =  \exp\left ( -\beta A (\beta) \right ) \times  \exp\left (C({\bf b}, \hat{w}) \right )  \label{eq:factorization}
 \ee  
 or that the total free energy must decompose into a sum of two terms
\be
F(\beta, {\bf b}, \hat{w}) \equiv - \frac{1}{\beta}\log {\cal Z}(\beta, {\bf b}, \hat{w}) =  A (\beta)  - \frac{1}{\beta}C({\bf b}, \hat{w}). \label{eq:decomposition}
\ee
The first term is a familiar thermodynamic free energy and, as we shall argue in the following section, the second term is related to a complexity of the neural networks. However, the free energy obtained from \eqref{eq:logZ} (with the local objectives parameter $m$ set for simplicity to zero) is
\be
F(\beta, {\bf b}, \hat{w}) = \frac{1}{2\beta} \sum_i \log \left ( 1   +  \beta \lambda_i \right )  -  \frac{N}{2 \beta} \log (2\pi)
\ee 
which does not in general decompose into a sum of two terms as in \eqref{eq:decomposition}. This suggest that in an equilibrium some additional restrictions must be imposed on the eigenvalues $\lambda_i$.  One possibility (that we shall verify numerically) is that 
\be
 \sum_{\lambda_i \gg  \beta^{-1}} \log \left ( 1 +  \beta \lambda_i \right ) \approx  \sum_{\lambda_i \gg \beta^{-1}}  \log(\lambda_i)   + {N_>} \log (\beta) \gg  \sum_{\lambda_i \lesssim  \beta^{-1}}  \beta \lambda_i \approx \sum_{\lambda_i \lesssim  \beta^{-1}} \log \left ( 1 +  \beta \lambda_i \right ) \notag
\ee
where $N_>$ is the number of eigenvalues $\lambda_i$ that are much greater than $\beta^{-1}$. Then the free energy can indeed be decomposed as in equation  \eqref{eq:decomposition},
\bea
F(\beta, {\bf b}, \hat{w}) &\approx& \frac{1}{2 \beta} \sum_{\lambda_i \gg \beta^{-1}}  \log(\lambda_i)   + \frac{N_>}{2\beta} \log (\beta) -  \frac{N}{2\beta} \log (2\pi) \label{eq:free}
\eea
with
\bea
 C({\bf b}, \hat{w})  &\approx& - \frac{1}{2}\sum_{\lambda_i \gg \beta^{-1}}  \log(\lambda_i) + \frac{N}{2} \log (2\pi)   \label{eq:complexity}
 \eea
 and
 \bea
 A(\beta) &\approx&  \frac{N_>}{2 \beta } \log (\beta)  \label{eq:A}.
\eea
Recall that  $\lambda_i$'s are the eigenvectors of $\hat{G}= \left ( \hat{I} - \hat{f}' \hat{w}\right)^T \left ( \hat{I} - \hat{f}' \hat{w}\right)$ and, thus, $\lambda_i$'s  are functions of ${\bf b}$ and $\hat{w}$.

\section{Thermodynamics of learning}\label{sec:thermodynamics}

The total Shannon entropy of the canonical ensemble can be obtained from the canonical partition function \eqref{eq:logZ},
\bea
S(\beta,{\bf b}, \hat{w} )  &=& -  \langle \log {p} \rangle =\beta^2  \frac{\partial}{\partial \beta } F(\beta, {\bf b}, \hat{w}).   \label{eq:total_entropy}
\eea
Just like the free energy, in a learning equilibrium \eqref{eq:equilibrium}, the entropy must also decompose into a sum of two terms,
\bea
S(\beta,{\bf b}, \hat{w} )  &=&  \beta^2  \frac{\partial}{\partial \beta } \left (  A (\beta)  - \frac{1}{\beta}C({\bf b}, \hat{w}) \right ) = \beta^2 \frac{\partial A(\beta)}{\partial \beta} +  C({\bf b}, \hat{w}). \label{eq:entropy_complexity}
\eea
The first term depends on only the inverse temperature parameter $\beta$ and we shall refer to it as a thermodynamic entropy
\be
{S}_0(\beta) =\beta^2 \frac{\partial A(\beta)}{\partial \beta}= \beta (U (\beta) - A(\beta) ).
\ee
For the total free energy \eqref{eq:free} it is given by
\bea
{S}_0(\beta)  &=&  -\beta A (\beta) + \beta U(\beta)\approx - \frac{N_>}{2} \log (\beta) + \frac{N_>}{2}  \notag\\
&\approx&   \frac{N_>}{2} \log (U) + \frac{N_>}{2} \left ( 1 - \log \frac{N_>}{2} \right )  \label{eq:thermodynamic_entropy}
\eea
where
\be
  U(\beta)  = -\frac{\partial}{\partial \beta} \log {\cal Z}(\beta, {\bf b}, \hat{w})  = \frac{\partial}{\partial \beta}\left ( \beta F(\beta, {\bf b}, \hat{w}) \right )  =  \frac{\partial}{\partial \beta}  \left (\beta A(\beta) \right ) \approx  \frac{N_>}{2\beta}.\label{eq:U}
\ee
 As the learning progresses, the average loss, $ U(\beta) $, decreases, the temperature parameter, $\beta^{-1}$, decreases and, thus, according to \eqref{eq:thermodynamic_entropy} one might expect that the thermodynamic entropy, ${S}_0$, should also decrease. However, it is not the thermodynamic entropy, ${S}_0$, but the total Shannon entropy $S$ (whose exponent describes accessible volume of the configuration space for ${\bf x}$) should become smaller with learning.  We shall call it the second law of learning (or perhaps the minus second law):\\
 \\
{\bf Second Law of Learning:} {\it the total entropy of a learning system can never increase during learning and is constant in a learning equilibrium,} 
 \be
 \frac{d}{dt} S  \le 0.\label{eq:second_law}
 \ee
 \\
 In the long run the system is expected to approach an equilibrium state with the smallest possible total entropy ${S}$ which corresponds to the lowest possible sum of the thermodynamic entropy, ${S}_0$, and of the complexity function $C({\bf b}, \hat{w})$ that we shall discuss next. 
 
In a feedforward neural network the weight matrix, $\hat{w}$, is nilpotent  \eqref{eq:nilpotent} and, therefore, the eigenvalues of the operator $\hat{w}$ are all zeros. This also implies that the eigenvalues of operator $ \hat{I} - \hat{f}' \hat{w}$ are all ones, but that does not tell us much about the eigenvalues of $\hat{G}$. On the other hand, the determinant of $\hat{G}$ is simply related to the determinant of  $ \hat{I} - \hat{f}' \hat{w}$, 
\be
\det \hat{G} = \det  \left ( \hat{I} - \hat{f}' \hat{w}\right)^2 = 1,
\ee
or
\be
\sum_i  \log(\lambda_i)  = 0. \label{eq:eig_log}
\ee
If we assume that near equilibrium $N_>$ does not change significantly, then a decrease in $C({\bf b}, \hat{w})$ implies that the largest eigenvalues   $\sum_{\lambda_i \gg \beta^{-1}}  \log(\lambda_i)$  of the operator $\hat{G}$  must increase and at the same time, according to \eqref{eq:eig_log}, the smallest eigenvalues $\sum_{\lambda_i \lesssim \beta^{-1}}  \log(\lambda_i)$ must decrease. Therefore, as the learning progresses, the operator $\hat{G}$ becomes better and better approximated by eigenvectors ${\bf v}_i$ with only largest eigenvalues, i.e.
\be
 \hat{G}  = \sum_{i} \lambda_i {\bf v}_i {\bf v}^T_i  =  \sum_{\lambda_i \gg \beta^{-1}} \lambda_i {\bf v}_i {\bf v}^T_i +  \sum_{\lambda_i \lesssim \beta^{-1}} \lambda_i {\bf v}_i {\bf v}^T_i  \approx \sum_{\lambda_i \gg \beta^{-1}} \lambda_i {\bf v}_i {\bf v}^T_i \label{eq:G_approx}
\ee 
This is what one might call a dynamical dimensional reduction of the state space (a subspace of dimension $N_> < N$ is sufficient to describe a state vector ${\bf x}$), or a reduction in the complexity of interconnections between neurons (a subspace of dimension $N_>^2  < N^2$ is sufficient to describe a weight matrix $\hat{w}$) or a complexity of computations that a given neural networks performs (a subspace of dimension $N_>^2 < N^2$  is sufficient to describe a linearized evolution operator $( \hat{I} - \hat{f}' \hat{w})$). For this reason we shall refer to $C({\bf b}, \hat{w})$ as a measure of complexity or just complexity. 

For a system transitioning between equilibrium states at constant temperature $T=1/\beta$, variations of the free energy must vanish, $d F=0$, and  then equation \eqref{eq:decomposition} takes the from of the first law,
\be
 d A  -  T d C = dU - T d S_0  -   T d C  = 0. \label{eq:first}
 \ee
or what we shall call the first law of learning (or perhaps the minus first law):
\\
\\
{\bf First Law of Learning:} {\it the increment in the loss function is proportional to the increment in the thermodynamic entropy  plus the increment   in the complexity} 
\be
dU =  T d S_0  +   T d C .\label{eq:first_law}
\ee
\\
This law describes how the learning system behaves when transitioning between equilibrium states, but in order to understand which neural architectures would be the most optimal we must take one step further and consider a non-equilibrium dynamics of the learning system. This is what we shall analyze in the following sections, but before continuing we would like to briefly discuss phenomenological aspects of the thermodynamic theory of learning. 

In this section we derived the first  \eqref{eq:first_law} and second  \eqref{eq:second_law} laws of learning for a system whose partition function is described by a Gaussian \eqref{eq:R2} in the limit when the total free energy consists of two terms \eqref{eq:free} which represent respectively the thermodynamic free energy \eqref{eq:A} and complexity of a neural network \eqref{eq:complexity}. To derive these results we had to make three assumptions. The first assumption is ergodicity, i.e. time average is the same as ensemble average. This is a rather standard assumption in the conventional statistical mechanics which can only be proved rigorously for very simple systems, but seems to work at a phenomenological level for a lot more general systems. We expect that in context of the learning dynamics a similar approach should work and under normal conditions the learning system can be treated at the level of statistical ensembles. The second  assumption is that the ensemble is Gaussian which allows us to calculate the partition function exactly, but for more general systems various interesting deviations from Gaussianity are expected. Once again, the situation is not very different from the conventional statistical mechanics where many perturbative and non-perturbative methods were already developed and these methods can be adopted for the analysis of the learning systems. And finally, the third assumption was that the total free energy is a sum of the thermodynamic free energy and complexity of a neural network.  This simple model provides a good description of the learning system that is analyzed numerically in Sec. \ref{sec:numerics}, but for more complicated systems the functional form of the total free energy would have to be modified. This is exactly what is usually done in the conventional thermodynamics where the shape of the total free energy is modeled phenomenologically based on the experimental data. Note that despite of the fact that the first and the second laws of thermodynamics (or in our case the first and second laws of learning) were derived for very simple thermodynamic systems (or in our case for learning systems), it is still expected that they would work for a more general systems given that the system is ergodic. However, there  are situations when the assumption of ergodicity is broken (e.g. in the glassy phase) and then some very interesting critical phenomena can emerge. We expect that the critical phenomena can also occur in the learning systems, but the analysis of such systems is beyond the scope of the present study.

\section{Optimal architecture}\label{sec:architecture}

Consider a family of bias vectors ${\bf b}({\bf Q})$ and weight matrices $\hat{w}({\bf Q})$ parametrized by dynamical parameters $Q_{k}$'s where ${k} \in (1,...,K)$. Typically the number of parameters $K$ is much smaller than  $N + N^2$ (i.e. the number of parameters required to describe a generic vector ${\bf b}$ and a generic matrix $\hat{w}$) and the art of designing  a neural architecture is to come up with functions  ${\bf b}({\bf Q})$ and $\hat{w}({\bf Q})$ which are most efficient in finding solutions. To make the statement more quantitative, consider an ensemble of neural networks described by a probability distribution  ${p}(\beta, {\bf Q})$ which evolves with ``time'' $\beta$ according to continuity equation
\be
\frac{\partial}{\partial \beta}{p}(\beta, {\bf Q}) = - \sum_{k}  \frac{\partial}{\partial Q_{{k}}}\left ( \frac{d Q_k}{d \beta} { p}(\beta, {\bf Q}) \right) \label{eq:dP}
\ee
where the parameters $Q_{k}$'s evolve in the direction of the gradient of the free energy
\be
\frac{d Q_{{k}}}{d \beta}  =  \alpha \frac{\partial F}{\partial Q_{k}}  =  \alpha \sum_{i} \frac{\partial b_{i}}{\partial Q_{k}} \frac{\partial F }{\partial b_{i}} + \alpha\sum_{i,j} \frac{d w_{ij}}{d Q_{k}} \frac{\partial F}{\partial w_{ij}}. \label{eq:dQ}
\ee
The Shannon entropy of the distribution ${p}(\beta, {\bf Q})$ with continuous variables ${\bf Q}$ (not to confuse with entropy $S(\beta,{\bf b}, \hat{w} )$ defined in the previous section) is, 
\be
{\mathscr S}(\beta) = - \int d^K Q \;\;{p}(\beta, {\bf Q}) \log \left ( {p}(\beta, {\bf Q}) \right ) \label{eq:entropy2}
\ee
Larger the entropy ${\mathscr S}(\beta)$, larger the accessible volume of the configuration space $\exp ({\mathscr S}(\beta))$, and therefore larger the rate with which new solutions for ${\bf b}$ and $\hat{w}$ can be found. Then an optimal architecture (described by ${\bf b}({\bf Q})$ and $\hat{w}({\bf Q})$) is the one for which the entropy destruction is minimized or, equivalently, the entropy production is maximized. We shall call it the principle of the minimum entropy destruction:
\\
\\
{\bf Principle of Minimum Entropy Destruction}: {\it The path taken by an optimal learning system is the one for which the entropy destruction is minimized (or the entropy production is maximized). }\\
\\
Note that the principle is the opposite of the minimum entropy production principle \cite{Prigogine,Klein} that is often used in context of non-equilibrium thermodynamics, but is consistent with the stationary entropy production principle that was recently used in context of emergent quantum mechanics \cite{entropic} and emergent gravity \cite{emergentgravity}.

In context of the learning systems, a useful expression for the entropy production can be obtained from \eqref{eq:dP}, \eqref{eq:dQ} and \eqref{eq:entropy2}, 
\bea
\frac{\partial }{\partial \beta }{\mathscr S}(\beta) & = &  - \frac{\partial }{\partial \beta } \int d^K Q \;\;{p} \log ({p})  =  - \int d^K Q \;\; \log ({p})  \frac{\partial }{\partial \beta } {p} \label{eq:entropy_change}\\
 & = &  \int d^K Q \;  \log ({p}) \sum_{k}  \frac{\partial}{\partial Q_{{k}}}\left ( \frac{d Q_k}{d \beta} { p}\right) = -  \int d^K Q \;  \sum_{k} \frac{d Q_k}{d \beta}  \frac{\partial {p}}{\partial Q_{{k}}} \notag\\
 & = &- \alpha \int d^K Q \;  \sum_{k} \frac{\partial F}{\partial Q_{k}}  \frac{\partial {p}}{\partial Q_{{k}}}  = \alpha \int d^K Q \;  \sum_{k} \frac{\partial^2 F}{\partial Q^2_{k}} {p} \notag \\
 & = &  \alpha \int d^K Q \;  {p}(\beta, {\bf Q}) \Delta F(\beta, {\bf Q})   \notag
\eea 
where the Laplacian operator is defined as usual $\Delta \equiv \sum_{k} \frac{\partial^2 }{\partial Q^2_{k}}$.

To better understand the optimization condition we can choose the parameters $Q_i$'s to be given by eigenvalues of the operator $\hat{G}$, i.e. $Q_i = \lambda_i$. Then the free energy \eqref{eq:logZ} (with $m$ set for simplicity to zero) can be approximated as
\bea
F  &\approx &  \frac{1}{2 \beta}  \log \det \left ( \hat{I} +  \beta\hat{G} \right )  \notag\\
&\approx & \frac{1}{2 \beta} \sum_i  \log  \left ( 1 +  \beta \lambda_i \right ) - \frac{N}{2\beta} \log (2\pi) \label{eq:free_eng}
\eea
 and its Laplacian as
 \bea
\Delta F&\approx & \frac{1}{2 \beta} \Delta \log \det \left ( \hat{I} +  \beta\hat{G} \right )  \notag\\
& \approx & \frac{1}{2 \beta} \sum_j \frac{\partial^2}{\partial \lambda_j^2} \sum_i  \log  \left ( 1 +  \beta \lambda_i \right ) \notag \\
&\approx & -\frac{\beta}{2} \sum_i \left ( 1+\beta \lambda_i \right)^{-2}. \label{eq:DeltaF}
 \eea
 Evidently, the Laplacian is always negative and, thus, for the entropy \eqref{eq:entropy_change} to decrease with learning $\alpha$ must be positive or that  the parameters $Q_{k}$'s evolve in the direction which maximizes the free energy \eqref{eq:dQ}. This may sound counterintuitive (since the main objective of learning is to minimize the loss function $U$), but it is a mere consequence of the decrease in the entropy $S$ which accompanies learning (i.e. the second law of learning).  Anyways, it should be clear that to improve learning efficiency we must choose an architecture such that the Laplacian $\Delta F$ is as close to zero as possible and then the entropy destruction is minimized (i.e. the principle of minimum entropy destruction). If we, once again, split all of the eigenvalues into large and small, then the Laplacian is approximately given by the number $N_<$  of smallest eigenvalues  $\lambda_i \ll \beta^{-1}$, i.e.
\be
\Delta F \approx -\frac{\beta}{2}  \sum_{\lambda_i \ll \beta^{-1}} \left ( 1+\beta \lambda_i \right)^{-2} \approx  -\frac{\beta}{2} N_<.
\ee
Note that while the largest eigenvalues $\lambda_i \gg \beta^{-1}$ are responsible for reducing complexity of the already obtained solutions \eqref{eq:complexity}, the  smallest eigenvalues $\lambda_i \ll \beta^{-1}$  are responsible for searching for new solutions.

 \section{Deep vs. shallow} \label{sec:deep_shallow}

\begin{figure}[]
\begin{center}
\includegraphics[width=1.0\textwidth]{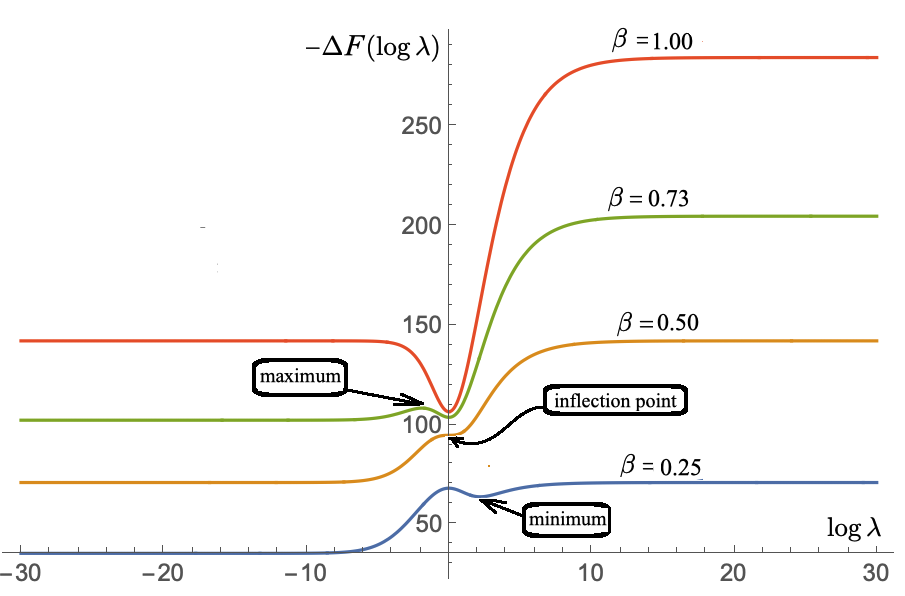}
\caption{$ -\Delta F(\log \lambda) $ for four different values of $\beta=0.25, 0.50, 0.73$ and $1.00$.} \label{fig:fig0}
\end{center}
\end{figure}

We are now ready  to tackle one of the biggest mysteries of machine learning. Why do deep neural networks perform so well? And we are not just asking why deep neural networks work at all (which is a well-known result known as the universal approximation theorem  \cite{Cybenko, Hornik}), but why and also when the neural networks should be efficient in finding solutions. We believe the answer is hidden in the free energy $ F$. As we have argued in the previous section \eqref{eq:entropy_change} the Laplacian $\Delta F$ describes the rate with which the entropy ${\mathscr S}$ decays and by minimizing, \be
- \Delta F= \frac{\beta}{2}  \sum_i \left ( 1+\beta \lambda_i \right)^{-2} =  \frac{\beta}{2}  Tr \left ( \hat{I} +\beta \hat{G} \right)^{-2}  \label{eq:min},
 \ee
 we maximize the efficiently of a neural network to find solutions (i.e. the principle of minimum entropy destruction).  To solve the minimization problem, we can consider a neural network with a small fraction  $\gamma \ll 1$ of eigenvalues  $\lambda_i \sim \lambda$ and a larger fraction $1-\gamma$  of  large eigenvalues at $\lambda_i \sim \lambda^{\frac{\gamma}{\gamma-1}}$ so that
\be
\det{\hat{G}} = \prod_i \lambda_i = \lambda^{\gamma N} \lambda^{\frac{\gamma}{\gamma-1} (1-\gamma) N} = 1.
\ee
Then the negative of the Laplacian is 
\bea
- \Delta F\approx  \frac{\beta}{2}  Tr \left ( \hat{I} +\beta \hat{G} \right)^{-2}     & = &\frac{N \beta}{2}  \left (\frac{ \gamma}{\left ( 1+\beta \lambda \right )^{2}}  + \frac{  1 - \gamma }{ \left ( 1+\beta \lambda^{\frac{\gamma}{\gamma-1}} \right )^{2} }\right ).
 \eea 
 On Fig. \ref{fig:fig0} we plotted $- \Delta F(\log \lambda)$ for four different values of the inverse temperature parameter $\beta=0.25, 0.50, 0.73, 1.00$, $\gamma=1/3$ and $N=854$. In the initial phase, $\beta < 1/2$, (e.g. blue line on Fig. \ref{fig:fig0}) there is a stable local minimum at $\log \lambda \approx 3 \log \left ( -\frac{1}{2} +  \sqrt{\frac{1}{4}+ \frac{1}{\beta}}  \right )$ and an unstable maximum at $ \log \lambda = 0$. In this phase, a small number, $N \gamma \ll N$, of eigenvalues is free to move away from a local maximum at $\log \lambda = 0$ to both smaller and larger values, but most of the eigenvalues $N (1-\gamma) \sim N$ should remain near $\log \lambda \sim 0$. In the intermediate phase, $1/2 < \beta < \sqrt{N} - 1 $, (e.g. green line on Fig. \ref{fig:fig0})  the two extreme points switch and there is an unstable maximum at 
\be
\log \lambda = 3 \log \left ( -\frac{1}{2} +  \sqrt{\frac{1}{4}+ \frac{1}{\beta}}  \right ). \label{eq:unstable_max}
\ee
In this phase only a decreasing fraction, $\gamma < (\beta + 1)^{-2}$, of the small eigenvalues, $\log \lambda < 3 \log \left ( -\frac{1}{2} +  \sqrt{\frac{1}{4}+ \frac{1}{\beta}}  \right )$, can still move to even smaller values, but the motion is terminated when the smallest values, $\log \lambda \ll 3 \log \left ( -\frac{1}{2} +  \sqrt{\frac{1}{4}+ \frac{1}{\beta}}  \right )$, reach the plateau. However, since $\det \hat{G} =1$, it is expected that the sum of the largest eigenvalues would continue to grow and according to \eqref{eq:complexity} the complexity of a network should continue to decrease. And finally, in the final phase, $ \beta >  \sqrt{N} - 1 $, (e.g. red line on Fig. \ref{fig:fig0})  the global minimum is at $\log \lambda =0$, the individual eigenvalues can no longer move towards $\log \lambda = - \infty$ and the ability of the neural network to learn becomes exponentially suppressed. 

This is what happens in an optimal system, however, by enforcing an architecture on a neural network (e.g. deep or shallow) we impose additional constraints on the  free energy (and on its Laplacian) which limits the ability of a network to explore the space of solutions. For example, in a feedforward neural network with many input neurons and few output/hidden neurons, most of the eigenvalues are set to $\log \lambda_i = 0$ and only a small fraction of eigenvalues is free to move to smaller and larger values. Clearly, the larger the number of the dynamical eigenvalues a neural architecture has, the better it is for learning. Therefore, in order to compare ``apples to apples'' we must first fix the number of the dynamical eigenvalues, and then look for an architecture which is flexible enough to support a skewed distribution of $\log \lambda_i$'s. As we have argued in the  previous paragraph, want we want is to be able to start with a single peak with all eigenvalues at $\log \lambda_i \sim 0$ and then to gradually grow a second peak with the largest eigenvalues  $\log \lambda_i \gtrsim 0  > - \log \beta$ which is to be balanced by the smallest eigenvalues $\log \lambda_i < - \log \beta$ that are dragged to smaller and smaller values.  With this respect, a better architecture is the one which supports a larger variance
\be
\mu_2 \equiv Tr \left ( \log \hat{G}  \right )^2 \label{eq:varience}
\ee
and a more skewed distribution or a more negative
\be
\mu_3 \equiv Tr \left ( \log \hat{G}  \right )^3. \label{eq:skew}
\ee
In a feedforward neural network the weight matrix is nilpotent \eqref{eq:nilpotent} as well as a product of the weight matrix, $\hat{w}$ and a diagonal matrix of first derivatives, $ \hat{f}'$, i.e. 
 \be
 \left( \hat{f}'\hat{w} \right )^n =  \left ( \hat{w}^T  \hat{f}'\right )^n = 0 \;\;\;\forall n \ge L \label{eq:powers}
 \ee
 where $L$ is the number of layers. For starters, consider a vary shallow network with no hidden layers (i.e. $L=2$) and thus $ \left ( \hat{w}^T  \hat{f}'\right )^2 = \left( \hat{f}'\hat{w} \right )^2=0$. Then there must exist functions $F_1(x)$ and $F_2(x)$ such that 
  \be
 \log \hat{G}  =  F_1(\hat{f}'\hat{w} \hat{w}^T \hat{f}')\,\hat{f}'\hat{w} + F_2( \hat{w}^T \hat{f}' \hat{f}'\hat{w}) \hat{w}^T \hat{f}' .
 \ee
and, therefore, 
\bea
Tr \left [  \log \hat{G}   \right ]  &=& 0\\
Tr \left [ \left ( \log \hat{G} \right )^2    \right ] &=& Tr \left [ F_2( \hat{w}^T \hat{f}' \hat{f}'\hat{w})  \hat{w}^T \hat{f}'  F_1(\hat{f}'\hat{w} \hat{w}^T \hat{f}')\,\hat{f}'\hat{w} + F_1(\hat{f}'\hat{w} \hat{w}^T \hat{f}')\,\hat{f}'\hat{w}F_2( \hat{w}^T \hat{f}' \hat{f}'\hat{w})  \hat{w}^T \hat{f}'    \right ]  \notag
\eea
and
\be
\mu_3 = Tr \left [  \left ( \log \hat{G} \right )^3 \right ] =  0.
\ee
In fact the traces of all odd powers must also be zero
\be
Tr \left [  \left ( \log \hat{G} \right )^{2n+1}  \right ]=  0
\ee
since every term in $ \left ( \log \hat{G} \right )^{2n+1}$ would have a product of unequal number of  $F_1(\hat{f}'\hat{w} \hat{w}^T \hat{f}')\,\hat{f}'\hat{w}$ and $F_2( \hat{w}^T \hat{f}' \hat{f}'\hat{w}) \hat{w}^T \hat{f}' $ terms which must be traceless. As we shall see in the next section, even with a single hidden layer (i.e. $L=3$) the second powers of operators $  \hat{f}' \hat{w}$ and $\hat{w}^T  \hat{f}'$ are very small and the skewness is still very small  $\mu_3 \approx  0$. What this means is that the effective number of dynamical eigenvalues is half of what it would have been if all eigenvalues were free to move without having to respect the symmetry of the distribution. However, as we keep adding more hidden layers the skewness of distribution grows larger, the eigenvalues become less constrained and the efficiency of learning is greatly improved. This might be why the deep learning is so efficient: hidden layers are essential for larger skewness $\mu_3$ and, as a result, for less negative Laplacian $\Delta F$ (and a slower decay of the entropy ${\mathscr S}$) which we claim is necessary for efficient learning. 
 
\section{Numerical experiments}\label{sec:numerics}

A direct numerical calculation of the distribution ${p}({\bf x}) $ is a computationally intensive task, but the main advantage of our statistical description is that the canonical ensemble \eqref{eq:canonical} can be viewed as purely phenomenological object. Then the main problem should be to come up with a model of the bulk loss function, $H({\bf x}, {\bf b}, \hat{w})$, which best describes the canonical ensemble and, consequently, the canonical partition function, ${\cal Z}(\beta, {\bf b}, \hat{w})$, and other thermodynamic quantities. On the other hand, the analysis of the preceding sections already suggests certain forms of the bulk loss function and of the partition function which we can easily verify numerically. In this section, we will check to what extent a feedforward neural network can be modeled by the bulk loss function without local objectives (i.e. \eqref{eq:bulk_loss} with $m=0$) or with the corresponding thermodynamic quantities: 

(a) average bulk loss (estimated in \eqref{eq:U}),
\be
 U(\beta)   =   \frac{N_>}{2 \beta} \label{eq:U_num},
 \ee
 
(b) complexity function (estimated in \eqref{eq:complexity}),
\be
 C({\bf b}, \hat{w})  = -\frac{1}{2} \sum_{\lambda_i \gg \beta^{-1}}  \log(\lambda_i) + \frac{N}{2} \log (2\pi) + \text{const} \label{eq:C_num},
\ee

(c) thermodynamic entropy (estimated in \eqref{eq:thermodynamic_entropy}),
\bea
 {S}_0(\beta) = -  \frac{N_>}{2} \log (\beta)  + \text{const}  =  \frac{N_>}{2} \log (U) - \frac{N_>}{2} \log \left ( \frac{N_>}{2}\right ) + \text{const}\label{eq:ent_num},
\eea
where $N_>$ is the number of eigenvalues $\lambda_i$'s much larger than $\beta^{-1}$. In addition, we will verify the expected dynamics of the eigenvalues and the anticipated dependence of the variance \eqref{eq:varience} and skewness \eqref{eq:skew} parameters on the performance of the neural networks obtained in the previous sections. 

All of the numerical experiments were carried out using the TensorFlow Python library  \cite{TensorFlow} and MNIST database of handwritten images \cite{MNIST}. Unfortunately, in the TensorFlow library the hidden layers are not dynamical and must be set prior to training. Nevertheless, we were able to obtain the desired results by running two different programs: the first one with two hidden layers (or what we shall call a ``deep'' neural network) and one with a single hidden layer (or what we shall call a ``shallow'' neural network).  In the deep network we used an input layer with $784$ neurons, the first hidden layer with $40$ neurons, the second hidden layer with $20$ neurons and the output layer with $10$ neurons; and in the shallow  network we used the same number of neurons on the input and output layers (i.e.  $784$ and $10$), but only a single hidden layer with $60$ neurons.  Altogether there are $N=784+40+20+10=784+60+10=854$ neurons in each neural network and so the state vectors $\bf x$ and the bias vectors $\bf b$ are $854$-dimensional vectors. The weight matrix $\hat{w}$ has  $854\times854$ components $w_{ij}$, but most of them are zero due to the predetermined architecture of hidden layers. The input layers represent a handwritten image of a number from $0$ to $9$ which is passed to $28\times28=784$ input neurons. One of the $10$ output neurons is to be activated only if the corresponding number is on the image. The activation function on all neurons is $f(y)=\tanh(y)$ and so the diagonal matrix $\hat{f}'$ has diagonal elements given by 
\be
f'_{ii}(y_i) =\frac{df(y_i)}{d y_i}=\frac{d\tanh(y_i)}{d y_i}= \sech(y_i)^2 = 4 (\exp(y_i)+\exp(-y_i))^{-2}.
\ee
The training was carried out using the method of stochastic gradient descent (SGD) for $30,000$ epochs with $6,000$  samples in the training dataset. We have also tried other optimizers such as Adam, but the results were essentially the same. 

\begin{figure}[]
\begin{center}
\includegraphics[width=0.8\textwidth]{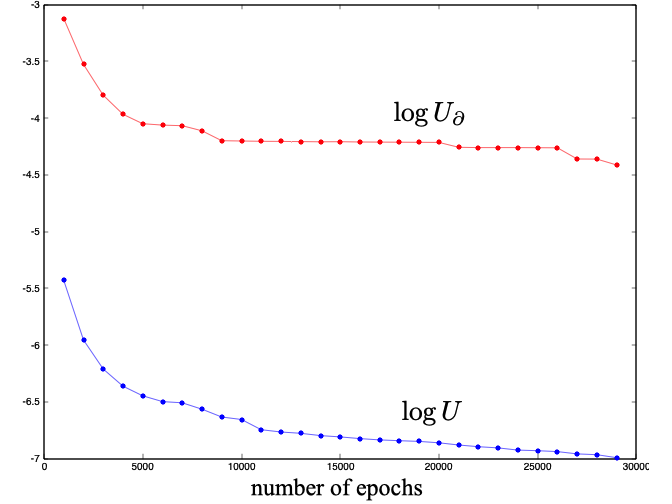}
\caption{Bulk loss (blue line) and boundary loss (red line) for $30,000$ training epochs.} \label{fig:fig1}
\end{center}
\end{figure}

\begin{figure}[]
\begin{center}
\includegraphics[width=0.8\textwidth]{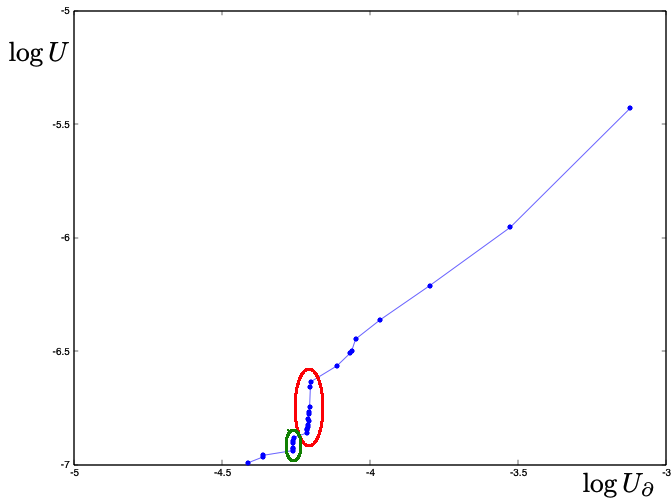}
\caption{Bulk loss vs. boundary loss for $30,000$ training epochs.} \label{fig:fig2}
\end{center}
\end{figure}

On Fig. \ref{fig:fig1} 
we plot (log of the ensemble-averaged) bulk loss $U= \langle H \rangle $ (blue line) and (log of the ensemble-averaged) boundary loss $U_\partial = \langle H_\partial \rangle $ from the deep neural network. As expected, the bulk loss remains few orders in magnitude smaller than the boundary loss, but both functions decrease with time. For training the neural network we used the (more familiar, but less general) boundary loss function $H_\partial$,  but a similar result is expected even if the (less familiar, but more general) bulk loss function $H$ would have been used instead. On Fig. \ref{fig:fig2} 
we plot the bulk loss $\log U $ vs.  the boundary loss $\log U_\partial $ from the same deep network. Note that at late times the bulk loss keeps decreasing while the boundary loss remains almost constant (inside of red oval on Fig. \ref{fig:fig2}). This behaviors continue for about $10,000$ (!) training epochs until the network finally finds a better solution and the boundary loss jumps to a smaller value (inside of green oval on Fig. \ref{fig:fig2}). And then essentially the same behavior continues, i.e. bulk loss decreases monotonically, but boundary loss makes sudden jumps. There is a simple explanation of the phenomena. The boundary loss is stuck in a saddle point with a large number of nearly flat directions for a very long time before it finds a way out. As the learning progress the system keeps moving along the flat directions and that does not reduce the boundary loss considerably, but the bulk loss and, as we shall see shortly, complexity keep decreasing with roughly the same pace. This shows that the bulk loss function has a lot fewer flat directions and with this respect a much better loss function. In addition, as we have argued in Sec. \ref{sec:unsupervised}, it  can be defined beyond supervised systems, e.g. for unsupervised learning. 

\begin{figure}[]
\begin{center}
\includegraphics[width=0.75\textwidth]{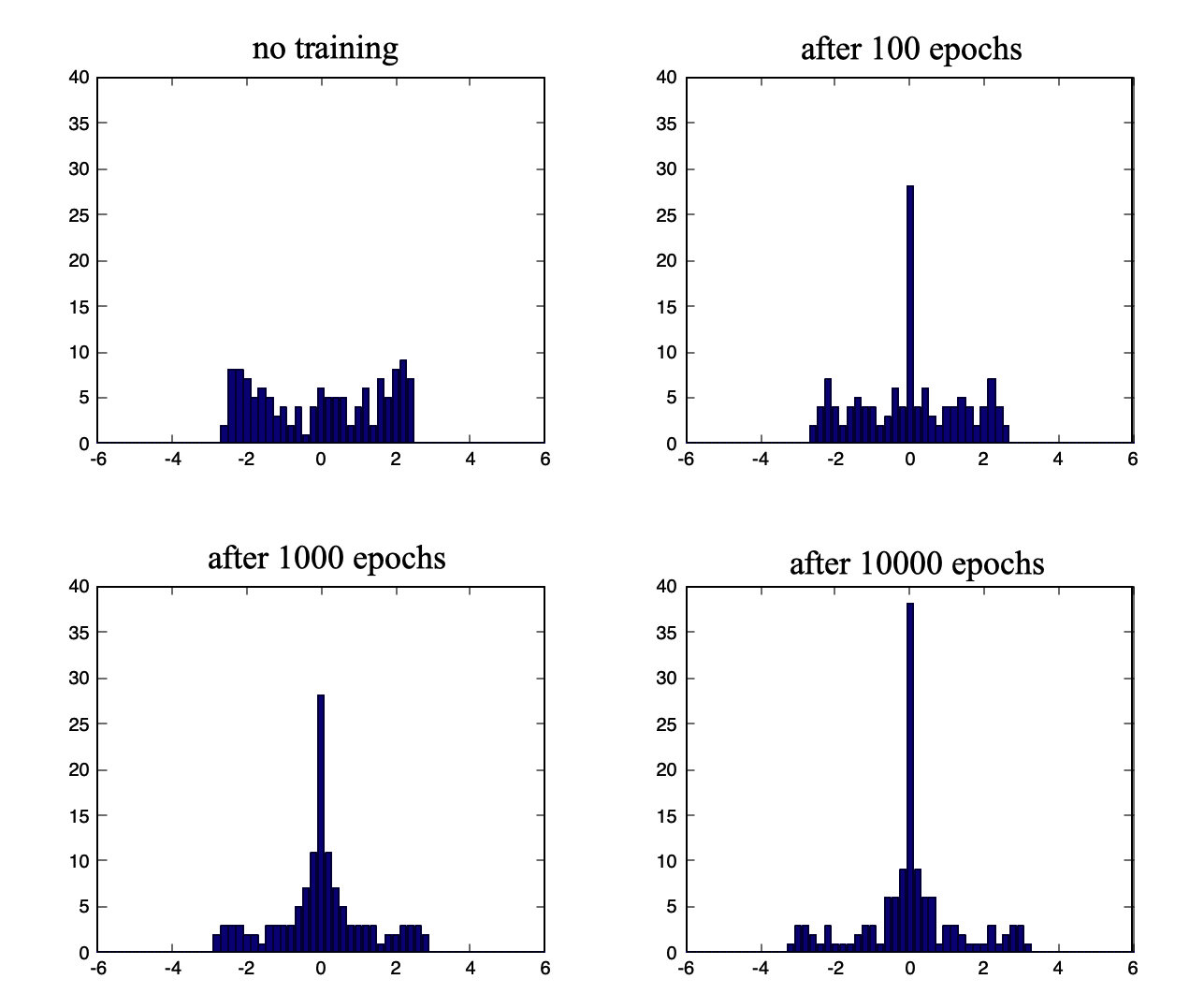}
\caption{ Histogram of eigenvalues of operator $\log \hat{G}$ from a shallow network.} \label{fig:fig3}
\includegraphics[width=0.75\textwidth]{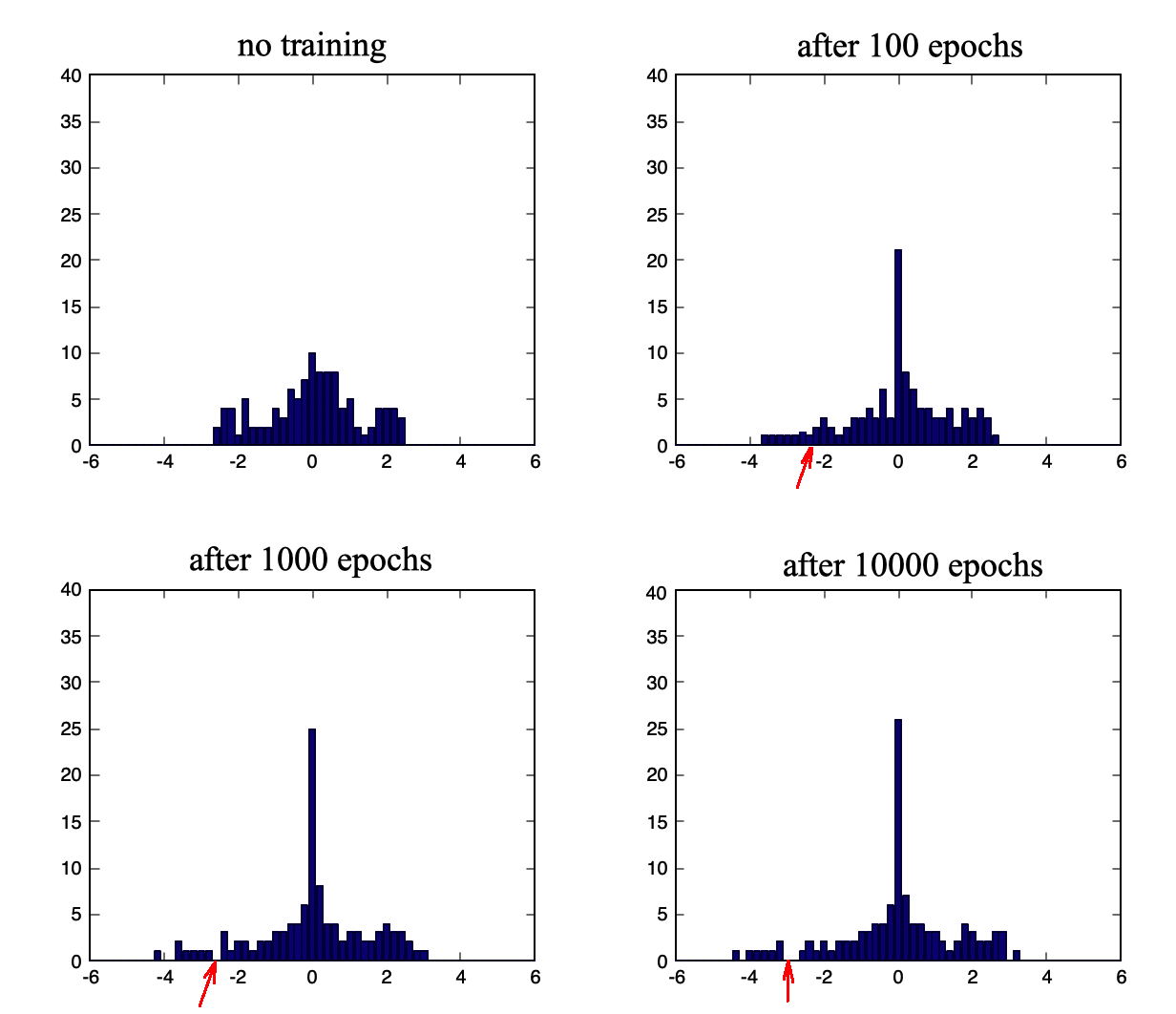}
\caption{ Histogram of eigenvalues of operator $\log \hat{G}$ from a deep network.} \label{fig:fig4}
\end{center}
\end{figure} 
On Figs. \ref{fig:fig3} and \ref{fig:fig4}
we plot histograms of the dynamical eigenvalues of operator $\log \hat{G}$ from the, respectively, shallow and deep networks. As expected, most of the eigenvalues remain near origin and only a fraction of eigenvalues is displaced significantly from $\log \lambda_i \sim 0$. The distribution of the eigenvalues in the shallow network is almost completely symmetric, the skewness after $10000$ epochs is  $\mu_3 \approx - 2.7 \times 10^{-10}$, the learning efficiency is suppressed and, as a result, the variance remains small $\mu_2\approx 1.84$. In contrast, the distribution of eigenvalues in the deep  network is not symmetric, the skewness after $10000$ epochs is more negative $\mu_3 \approx -1.7$, the learning efficiency is enhanced and the variance grows larger $\mu_2\approx 2.44$. There is a clear gap between the smallest and larger eigenvalues (marked by red arrows on Fig. \ref{fig:fig4}) which can be seen after  $10000$ epochs at $\log \lambda \approx -3.0$, after  $1000$ epochs at $\log \lambda \approx -2.5$ and may be even after $100$ epochs at $\log \lambda \approx -2.3$. This gap is expected to be at unstable maximum defined by equation \eqref{eq:unstable_max} which implies that after $100$ epochs $\beta \approx 1.37$,  after $1000$ epochs $\beta \approx 1.40$  and  after $10000$ epochs $\beta \approx 1.47$. In the shallow network the gap cannot be clearly identified since it is closer to the origin and the inverse temperature parameter $\beta$ is smaller.  Also note that while the smallest eigenvalues move to smaller values, to satisfy \eqref{eq:eig_log} the largest eigenvalues must move to larger values. Recall, that the largest eigenvalues describe the complexity of the network \eqref{eq:C_num} and the increase of the largest eigenvalues represents a decrease in the complexity of the network. 
 
\begin{figure}[]
\begin{center}
\includegraphics[width=0.81\textwidth]{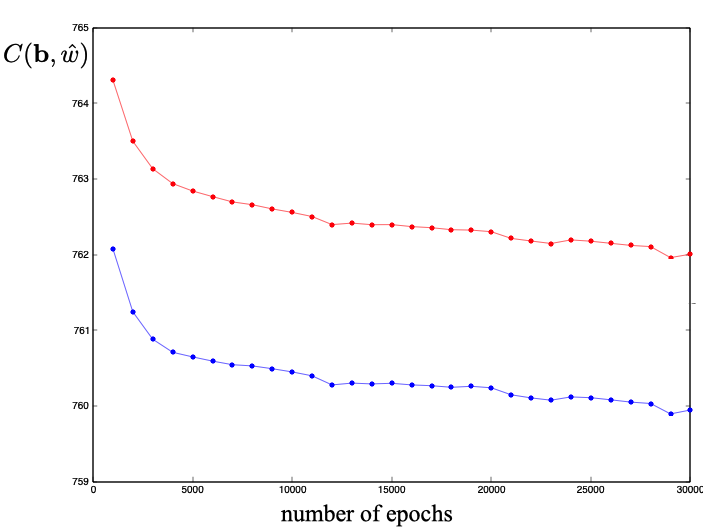}
\caption{ Complexity $C({\bf b}, \hat{w})$ of a deep neural network as a function of the number of training epochs.} \label{fig:fig5}
\includegraphics[width=0.81\textwidth]{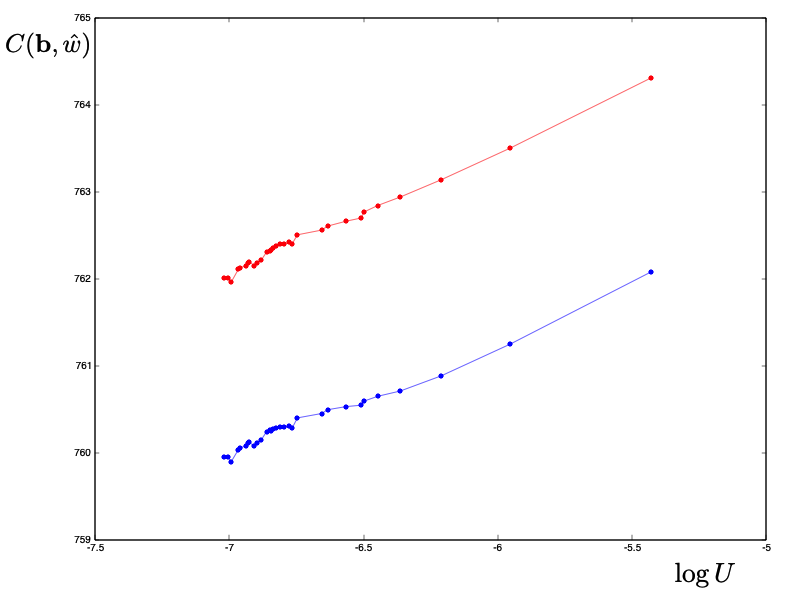}
\caption{ Complexity $C({\bf b}, \hat{w})$ of a deep neural network as a function of the bulk loss $\log U$.} \label{fig:fig6}
\end{center}
\end{figure}

In the previous paragraph, we estimated the values of $\beta$ by identifying a gap (marked by red arrows on Figs. \ref{fig:fig4}) between the smallest eigenvalues and the rest. However, as one can see from Fig. \ref{fig:fig4} the smallest eigenvalues $\log \lambda_i < -\log \beta$ keep moving to smaller values together with $-\log \beta$. This suggests that (instead of using equation \eqref{eq:C_num}) we can try to define an approximate complexity by a sum of a fixed number of the largest eigenvalues,
\be
 C_{n}({\bf b}, \hat{w})  = -\frac{1}{2} \sum_{i = 1}^n  \log(\lambda_i) + \frac{N}{2} \log (2\pi),
\ee
where it is assumed that the eigenvalues are ordered $\lambda_1 \ge \lambda_2 \ge ... \ge \lambda_N$. On Fig. \ref{fig:fig5}
we plot two (limiting) complexities by summing over twenty largest eigenvalues,  $C_{20}({\bf b}, \hat{w})$, (blue line) and by summing over all but twenty smallest eigenvalues, $C_{834}({\bf b}, \hat{w})$, (red line). Evidently, up to an additive constant, the behavior of both curves is similar and so either one (or anyone in-between) can be used to study a relaxation of the system towards equilibrium. On Fig. \ref{fig:fig6} we plot the same complexities, but as a function of the bulk loss $\log U$. Both functions are nearly linear with slopes of order one:  $C_{20}({\bf b}, \hat{w}) \approx 769+ 1.25 \log U$ for the blue line and $C_{834}({\bf b}, \hat{w}) \approx 772 + 1.41 \log U$ for the red line. In fact the linear dependance is in agreement with the second law of learning \eqref{eq:second_law}  which states that the total entropy must decay with learning. Recall the total entropy is a sum of the complexity \eqref{eq:C_num}  and thermodynamic entropy \eqref{eq:ent_num} (which scales linearly with $\log U$ whenever $N_>$ remains constant) and so it is expected that both quantities would decay with roughly the same rate. 

There are certainly many other numerical experiments that we could have done, but it should already be evident that the statistical description developed in the paper might actually shed light on what is happening behind scenes in machine learning. We now switch to the most speculative part of the paper by asking if the entire universe on its most fundamental level could be described by a neural network. In the following sections, we shall only describe the main idea and an interested reader is referred to Refs. \cite{world, quantumness} for a more in-depth analysis of the proposal.  

\section{Entropic mechanics}\label{sec:entropic}

Quantum mechanics is a remarkably successful paradigm for modeling physical phenomena observed on a wide range of scales ranging from  $10^{-19}$ meters (i.e. high-energy experiments, e.g. \cite{PDG}) to $10^{+26}$ meters (i.e. cosmological observations, e.g. \cite{Planck}.) The paradigm is so successful that it is widely believed that on the most fundamental level the entire universe is governed by the rules of quantum mechanics and even gravity should somehow emerge from it. This is known as the problem of quantum gravity that so far has not been solved, but some progress has been made in context of AdS/CFT correspondence \cite{Maldacena, Witten, Susskind}, emergent gravity \cite{Jacobson, Verlinde, emergentgravity}, quantum entanglement \cite{Ryu, Swingle,Almheiri}  and holographic complexity \cite{Complexity,  Complexity1, Complexity2}.\footnote{There seems to be an interesting connection between thermodynamics of learning systems (see Sec. \ref{sec:thermodynamics}) and thermodynamics of holographic complexity. In Ref. \cite{Complexity1} the authors showed that the quantum computational complexity of a holographic states on the anti-de Sitter boundary is dual to an action over a Wheeler-de Witt patch in the bulk. In the learning systems, the complexity function $C({\bf b}, \hat{w})$ is the quantity which best describes the complexity of a boundary state (for example, in a feedforward network $C({\bf b}, \hat{w})$ is the complexity of a neural network which maps the input boundary data to output boundary data with the smallest error) and a thermodynamic free energy $A(\beta)$ is the quantity which best describes the state of the local degrees of freedom in the bulk. According to the First Law of learning \eqref{eq:first_law} the two quantities are in fact related if not in an absolute sense, then at least in a relative sense, i.e.
\be
d A  = dU - T d S_0 =  T d C.
\ee
Also note that even the Second Law of learning \eqref{eq:second_law} is somewhat related to the recently proposed Second Law of complexity of quantum states \cite{Complexity2}  with the main difference that during learning the complexity $C$ decreases not on its own, but together with the thermodynamic entropy $S_0$.
This suggests that there might be a deep connection between learning systems and holographic systems that we still have not figured out.} Although extremely important, the problem of quantum gravity is not the only problem with quantum mechanics. The quantum framework also starts to fall apart with introduction of observers. Everything seems to work very well when observers are kept outside of a quantum system, but it is far less clear how to described macroscopic observers in a quantum system such as the universe itself. The realization of the problem triggered an ongoing debate on the interpretations of quantum mechanics, which remains unsettled to this day. On one side of the debate, there is an increasing number of proponents of the many-worlds interpretation claiming that everything in the universe (including observers) must be governed by the Schr\"odinger equation \cite{Everett}, but then it is not clear how classical probabilities would emerge. One the other side of the debate, there are proponents of the hidden variables theories \cite{Bohm}, but there it is also unclear what is the role of the wave-function in a purely statistical system. It is important to emphasize that a working definition of observers is necessary not only for settling some philosophical debates, but for understanding the results of real physical experiments and cosmological observations. In particular, a self-content, paradoxes-free definition of observers would allow us to understand the significance of  Bell's inequalities \cite{Bell} and to make probabilistic prediction in cosmology \cite{Measure}. In the absence of such definition, the so-called Copenhagen interpretation is often adopted, which is based on a clear cut between the observer and the system, modeled using quantum mechanics \cite{Bohr}. This works very well in context of, for example, condensed matter and high energy experiments, where the observer remains outside of the system, but is not satisfactory in context of cosmology and quantum gravity, where observers are essential parts of the system. 

To resolve the apparent inconsistency  (or incompleteness) in our description of the physical world, we shall entertain a (not so new) idea of having a more fundamental theory than quantum mechanics. A working hypothesis is that on the most fundamental level the dynamics of the entire universe is described by a microscopic neural network. If correct, then not only macroscopic observers should emerge from the microscopic neural network  (see, for example, \cite{Tononi}), but, more importantly, the equations of quantum mechanics and general relativity should correctly describe an emergent dynamics of the corresponding learning system. Our main goal in this section is to show that quantum mechanics (or, more precisely, Schr\"odinger equation) indeed provides a good description of an optimal neural network not too far from an equilibrium and we postpone the discussion of general relativity until the next section. Note that most of the results in the remainder of this section were originally obtained in Ref. \cite{entropic}, but in a slightly different context and with slightly different assumptions. 

Recall that equation \eqref{eq:dP} describes evolution of a probability distribution ${p}(\beta, {\bf Q})$ where $Q_{k}$'s (for ${k} \in (1,...,K)$)  parametrize the weight matrix $\hat{w}({\bf Q})$ and the bias vector ${\bf b}({\bf Q})$. The equation works well away from an equilibrium, but at an equilibrium the first derivatives of the free energy remain small
\be
\frac{d Q_{{k}}}{d \beta}  =  \alpha \frac{\partial F}{\partial Q_{k}}
\ee
and the dominant contribution to the entropy production comes from ``diffusion''. Then we can study evolution of the system along the equilibrium manifold of dimension $K - \tilde{K} \ll  K $ (i.e. the number of  ``Goldstone'' modes is $K - \tilde{K}$) defined by a (degenerate) maximum of the free energy $F( {\bf Q})$. More precisely, the equilibrium manifold can be described by 
\bea
\frac{d F(t, {\bf Q})}{dt} =  \frac{\partial F(t, {\bf Q})}{\partial t}  + \sum _k \frac{d Q_{{k}}}{d t} \frac{\partial F(t, {\bf Q})}{\partial Q_k} \notag\\
=  \frac{\partial F(t, {\bf Q})}{\partial t}  +\alpha \sum _k \left ( \frac{\partial F(t, {\bf Q})}{\partial Q_k} \right )^2 = 0 \label{eq:F_const}
\eea
where the first term represents the change of the free energy due to dynamics of $x_i$'s and the second term represents the change in the free energy due to dynamics of $Q_k$'s. Note that instead of studying the dynamics in $\beta$ we switched to a new parameter $t$ (e.g. the number of training epochs) for which the dynamics on small time scales can be approximated by a Fokker-Planck equation
\be
\frac{\partial p(t, {\bf Q})}{\partial t}\approx \frac{D}{2} \sum_{{{k}}} \frac{\partial^2 p(t, {\bf Q})}{\partial {Q}^2_{{{k}}}} = \frac{D}{2} \Delta {p}(t, {\bf Q}) \label{eq:dP3}
\ee
with a time-independent diffusions coefficient $D$. However, on longer time scales the diffusion must be constrained to satisfy \eqref{eq:F_const}, but in our statistical description we shall only demand that it is satisfied on average, i.e.
\be
\int d^K Q\;  {p}(t, {\bf Q}) \left (  \frac{\partial F(t, {\bf Q})}{\partial t}  +\alpha \sum _k \left ( \frac{\partial F(t, {\bf Q})}{\partial Q_k} \right )^2\right )  = 0. \label{eq:constraint}
\ee
 For simplicity, we also assume that the diffusion coefficient $D$ does not depend on $\bf Q$ and so no factor ordering problems arise. Then the main problem is to solve for ${p}(t, {\bf Q})$ subject to constraint \eqref{eq:constraint} which is exactly the type of problems considered in Ref. \cite{entropic}. There it was shown that a constrained dynamics can often be described by an approximate Schr\"odinger equation if the so-called principle of stationary entropy production is satisfied:
\\
\\
{\bf Principle of Stationary Entropy Production}: {\it The path taken by the system is the one for which the entropy production is stationary. }\\
\\
However, in Sec. \ref{sec:architecture} we argued that in an optimal architecture a closely related principle (of minimum entropy destruction) should be satisfied and, therefore, all that we need to assume is that the microscopic neural network has an optimal architecture. 

The optimization problem can then be solved by combining into a single functional ${\cal S}(p, F) $ two terms: the total entropy production from time $t=0$ to time $t=T$ and constraint \eqref{eq:constraint} imposed by on the time-dependent free energy, i.e.
\bea
{\cal S}(p, F)&=& - \int_0^T dt  \frac{d }{d t} \int d^K Q \;\;{p} \log ( {p})   + \int_0^T dt  \int d^K Q \, {p} \left (  \frac{\partial F}{\partial t}  +\alpha \sum _k \left ( \frac{\partial F}{\partial Q_k} \right )^2\right )  \notag\\
&=&  \int_0^T dt \int d^K Q \;\; \left ( - \log ( {p})  \frac{D}{2} \sum_{{{k}}} \frac{\partial^2 p}{\partial {Q}^2_{{{k}}}}     + {p} \left (  \frac{\partial F}{\partial t}  +\alpha \sum _k \left ( \frac{\partial F}{\partial Q_k} \right )^2\right )  \right ).\label{eq:entropy_production}
\eea
After integrating by parts and ignoring the boundary terms (assuming that $p$ vanishes at the boundaries of integration) we obtain
\bea
{\cal S}(p, F)&=&  \int_0^T dt \int d^K Q \;\; \left (  2 D \sum_{{{k}}}  \left (  \frac{\partial \sqrt{p}  }{\partial {Q}_{{{k}}}}  \right )^2   + {p} \left (  \frac{\partial F}{\partial t}  +\alpha \sum _k \left ( \frac{\partial F}{\partial Q_k} \right )^2\right )  \right ) \notag\\ 
&=&  \int_0^T dt \int d^K Q \;\; \sqrt{p}  \left ( -  2 D \sum_{{{k}}} \frac{\partial^2 }{\partial {Q}^2_{{{k}}}} +\alpha \sum _k \left ( \frac{\partial F}{\partial Q_k} \right )^2   +  \frac{\partial F}{\partial t}  \right )  \sqrt{p} \notag\\ 
&=&   \int_0^T dt \int d^K Q \;\; \Psi^*   \left (  - 2 D\Delta  - i  \sqrt{\frac{2D}{\alpha}} \frac{d}{d t} \right )  \Psi  \label{eq:Lagrangian}
\eea
where the wave-function is defined as
\be
\Psi (t, {\bf Q}) \equiv \sqrt{p (t, {\bf Q})}  \exp \left (i \sqrt{\frac{\alpha}{2 D}} F(t, {\bf Q}) \right ).
\ee
Evidently, upon varying \eqref{eq:Lagrangian} we arrive at a Schr\"odinger equation 
\be
-  i \frac{d}{dt } \Psi (t, {\bf Q}) = \sqrt{ 2D \alpha}\; \Delta \Psi (t, {\bf Q})
\ee
whose solutions extremize the functional ${\cal S}(p, F)$ or, in other words, describe a trajectory in the configuration space which minimizes entropy destruction. (See Ref. \cite{entropic} for details). Therefore, we conclude that quantum mechanics (or at least Schr\"odinger equation) can in fact emerge from a microscopic neural network with an optimal architecture near equilibrium.

\section{Emergent gravity}\label{sec:gravity} 

Now we turn to gravity.\footnote{As far as we know the only attempt to describe gravity in terms of quantum neural networks was made in Ref. \cite{Dvali}. However, the main difference with our approach is that the microscopic neural network considered here is not quantum, but statistical. On the other hand, as we have argued in Sec. \ref{sec:entropic}, the quantum behavior of the microscopic neural network is expected and so it is possible that the two systems are equivalent.} If the microscopic neural network has an optimal architecture then it is still the case that the principle of minimum entropy destruction (or, the more general, principle of stationary entropy production) should be satisfied and so the relevant quantity to extremize is still \eqref{eq:entropy_production}, which contains both entropy production (first term) and constraints (second term). What is, however, different is that we must allow for the larger system to be further away from a learning equilibrium and so the number of constraints can be much smaller than the number of parameters $K$. In other words, the dimensionality of the equilibrium manifold (or, if you wish, the number of symmetries) remains very high. This implies that the probability distribution $p(t, {\bf Q})$ should have a higher degree of symmetry and thus can be parametrized $p(\hat{g}, {\bf Q})$ with a (relatively) small number of auxiliary parameters $\hat{g}(t)$. For example, if the probability distribution is parametrized by Gaussian distributions, $p(\hat{g}(t), {\bf Q}) \propto  \exp\left (- {\sum_{k,k'} g_{k k'}(t) Q_k Q_{k'} } \right )$, then the optimization problem is to find  $\hat{g}(t)$ and ${\bf \Phi}(t, {\bf Q})$ which extremize \eqref{eq:entropy_production}, 
\be
{\cal S}(\hat{g}, {\bf \Phi})= \int_0^T dt  \sum_k  \left  \langle - \frac{D}{2} \frac{\partial^2 \log p(\hat{g}(t), {\bf Q})}{\partial {Q}^2_{{{k}}}}  \right \rangle  +  \int_0^T dt  \left \langle {\cal L}(\hat{g}(t), {\bf \Phi}(t, {\bf Q}) \right \rangle.\label{eq:Lagrangian2}
\ee 
The first term represents the entropy production and depends only on $\hat{g}$ and the second term represents constraints and depends on both $\hat{g}$ and some parameters ${\bf \Phi}(t, {\bf Q})$ which enforce the constraints. (For example, in the previous section ${\bf \Phi}(t, {\bf Q})$ was given by  free energy $F(t, {\bf Q})$ which is to be extremized in an equilibrium). 

This is what might be happening on a microscopic level, but our task in this section is to only develop a phenomenological model of gravity based on what we already know about general relativity. In gravitational theories the dynamical degrees of freedom are described by a metric tensor $g_{\mu\nu}(x)$ and other fields $\varPhi(x)$ all of which are functions of four space-time coordinates $x = (x^0, x^1, x^2, x^3)$. From that perspective a better parametrization of the probability distribution is given by $g_{\mu\nu}(x)$ and of the constraints by $\varPhi(x)$. Then \eqref{eq:Lagrangian2} can be expressed phenomenologically as 
\be
{\cal S}(g_{\mu\nu}, \varPhi) = \int d^{D+1} x\;\sqrt{|g|} \; \left ( - \frac{1}{2\kappa}{R}({g}_{\mu\nu}(x)) + \Lambda \right ) + \int d^{D+1} x\;\sqrt{|g|} \; {\cal L}(g_{\mu\nu}(x), \varPhi(x))  \label{eq:full_action}
\ee
where, as before, the first term represents the entropy production and the second term represents the constraints. Several comments are in order. First of all, in equation \eqref{eq:Lagrangian2} the parameter $\hat{g}$ was a finite dimensional matrix, but  in equation \eqref{eq:full_action} the parameter $g_{\mu\nu}(x)$ is a continuous function and so at best it is an approximate mapping which should break down at some UV scale (e.g. Planck scale). Secondly, even if the metric tensor  $g_{\mu\nu}(x)$ is defined only on some very fine-grained  lattice, there is a sense of distance between gravitational degrees of freedom which is not present in a neural network. This would be true for a general learning system, but we expect that for a clever choice of local objectives the weight matrix $\hat{w}$ (which is also an adjacency matrix describing the strength of connections between neurons) could be attracted towards a three-dimensional lattice (see  \cite{graphflow} for a possible mechanism) and then the space-time locality would emerge. Thirdly, any lattice-like structure would break a general covariance which is known to be a very precise symmetry of nature. Therefore, we must also assume that the local objectives of neurons are such that the general covariance would emerge on large scales (see  \cite{spacetime} for a possible mechanism), but exactly  how this might work is presently unknown.

In the remainder of this section we shall follow closely a phenomenological procedure outline in Ref. \cite{emergentgravity}. We first expand the entropy production term in \eqref{eq:full_action} around equilibrium, i.e.
\be
\frac{1}{2\kappa}{R} = g_{\alpha\beta, \mu} J^{\mu\alpha\beta} 
\ee
where the fluxes are denoted by $J^{{k}\alpha\beta}$ and the generalized forces are taken to be\footnote{Summations over repeated indices are implied everywhere in this section.}
\be
g_{\alpha\beta, \mu } \equiv \frac{\partial g_{\alpha\beta} }{\partial x^{\mu }}.
\ee
Then we can expand fluxes around local equilibrium to the linear order in generalized forces
\be
J^{\mu\alpha\beta} =  L^{\mu\nu\;\alpha\beta\;\gamma\delta} g_{\gamma\delta, \nu} .
\ee
to obtain 
\be
\frac{1}{2\kappa}{R} = L^{\mu\nu\;\alpha\beta\;\gamma\delta} g_{\alpha\beta, \mu}  g_{\gamma\delta, \nu } \label{eq:Onsager2}.
\ee
One can think of \eqref{eq:Onsager2} as a defining equation for the Onsager tensor, but then we are forced to only consider Onsager tensors $ L^{\mu\nu\;\alpha\beta\;\gamma\delta}$ that are symmetric under interchanges   $(\mu, \alpha,\beta) \leftrightarrow (\nu, \gamma, \delta)$, i.e.
 \be
 L^{\mu\nu\;\alpha\beta\;\gamma\delta} = L^{\nu\mu \;\beta\alpha\;\delta\gamma}.  \label{eq:onsager_relation}
 \ee
These are the Onsager reciprocity relations \cite{Onsager} for our learning system, but there are also other (trivial) symmetries that one should impose $(\alpha) \leftrightarrow (\beta)$, $(\gamma) \leftrightarrow (\delta)$, due to symmetries of the metric, i.e.
\be
 L^{\mu\nu\;\alpha\beta\;\gamma\delta}  =  L^{\mu\nu\;(\alpha\beta)\;(\gamma\delta)}. \label{eq:correct_choice} 
\ee
The overall space of such tensors is still pretty large, but it turns out that a very simple choice leads to general relativity:
\be
L^{\mu\nu\;\alpha\beta\;\gamma\delta} =  \frac{1}{8\kappa} \left (  g^{\alpha\nu} g^{\beta\delta} g^{\mu\gamma} + g^{\alpha\gamma} g^{\beta\nu}g^{\mu\delta} - g^{\alpha\gamma} g^{\beta\delta} g^{\mu\nu} - g^{\alpha\beta} g^{\gamma\delta} g^{\mu\nu} \right ) .\label{eq:Onsager}
\ee
After integrating by parts, neglecting boundary terms and collecting all other terms we get
\bea
\int d^{D+1} x &\sqrt{|g|} &\frac{1}{2\kappa}{R}  =
  \int d^{D+1} x \sqrt{|g|} g^{\mu\nu} \;\; \frac{1}{\kappa} \left (\Gamma^{\alpha}_{\phantom{\alpha}\nu[\mu, \alpha]} +  \Gamma^{\beta}_{\phantom{\beta}\nu[\mu} \Gamma^\alpha_{\phantom{\alpha}\alpha]\beta}\right )=\\\notag  
  &=&\int d^{D+1} x \sqrt{|g|} \;\; \frac{1}{8\kappa} \left (  g^{\alpha\nu} g^{\beta\delta} g^{\mu\gamma} +g^{\alpha\gamma} g^{\beta\nu} g^{\mu\delta} -g^{\alpha\gamma} g^{\beta\delta} g^{\mu\nu} -g^{\alpha\beta} g^{\gamma\delta} g^{\mu\nu} \right )  g_{\alpha\beta,\mu} g_{\gamma\delta,\nu}\;
\eea
 where
 \bea
 \Gamma^{\mu}_{\phantom{\mu}\gamma\delta} \equiv \frac{1}{2} g^{\mu\nu} \left ( g_{\nu\gamma,\delta} +  g_{\nu\delta,\gamma} -  g_{\gamma\delta, \nu} \right ) \;\;\;\;\;\;\;\;\;\;\text{and}\;\;\;\;\;\;\;\;
 \Gamma^\alpha_{\phantom{\alpha}\mu\nu, \beta} \equiv \frac{\partial}{\partial x^\beta}  \Gamma^\alpha_{\phantom{\alpha}\mu\nu}.
 \eea
Therefore, upon varying \eqref{eq:full_action} with respect to the metric $g^{\mu\nu}$ we get the Einstein equations
\be
R_{\mu\nu} - \frac{1}{2} Rg_{\mu\nu} + \Lambda g_{\mu\nu} = \kappa  T_{\mu\nu} 
\ee
where the Ricci tensor is 
\be
R_{\mu\nu} \equiv 2 \left (\Gamma^{\alpha}_{\phantom{\alpha}\nu[\mu, \alpha]} +  \Gamma^{\beta}_{\phantom{\beta}\nu[\mu} \Gamma^\alpha_{\phantom{\alpha}\alpha]\beta}\right )
\ee
and the energy-momentum tensor is 
\be
 T_{\mu\nu} \equiv -\frac{2}{\sqrt{|g|}} \frac{\delta (\sqrt{|g|} {\cal L})}{\delta g^{\mu\nu}}.
\ee
(See Ref. \cite{emergentgravity} for details). Of course, the expectations are that this result would only hold near equilibrium, and there should be deviations from general relativity when some of the symmetries in the Onsager tensor \eqref{eq:Onsager} are broken.  \\

{\it Acknowledgments.}  I would like to express my sincere  gratitude to my former teacher, Walter Johnson, who introduced me to the subject of artificial neural networks. This work was supported in part by the Foundational Questions Institute (FQXi).


\begin{thebibliography}{10}

\bibitem{Saxe} A.~M.~Saxe, J.~L.~McClelland, S.~Ganguli, ``Exact solutions to the nonlinear dynamics of learning in deep linear neural networks,'' In the International Conference on Learning Representations, (2014)

\bibitem{Choromanska} A.~Choromanska, M.~Henaff, M.~Mathieu, G.~Arous, Y.~LeCun, ``The Loss Surfaces of Multilayer Networks,'' In Proceedings of the 18th International Conference on Artificial Intelligence, volume 38, (2015)

\bibitem{Kadmon} J.~Kadmon, H.~Sompolinsky, ``Optimal Architectures in a Solvable Model of Deep Networks,'' In Advances in Neural Information Processing Systems, (2016)

\bibitem{Bottleneck2}
R.~ Shwartz-Ziv, N.~Tishby, ``Opening the black box of deep neural networks via information,'' arXiv:1703.00810 [cs.LG], (2017)

\bibitem{Advani} M.~S. ~Advani, A.~M.~Saxe. ``High-dimensional dynamics of generalization error in neural networks,'' arXiv preprint arXiv:1710.03667 (2017)

\bibitem{LinTegmarkRolnick}
H.~W.~Lin,  M.~Tegmark, D.~Rolnick, ``Why Does Deep and Cheap Learning Work So Well?'', Journal of Statistical Physics, Volume 168, Issue 6, pp.1223-1247 (2017)

\bibitem{Galushkin} A.I. Galushkin, ``Neural Networks Theory,'' Springer, 396 p., (2007)

\bibitem{Schmidhuber}  J.~Schmidhuber, ``Deep Learning in Neural Networks: An Overview,'' Neural Networks. 61: 85-117. (2015)

\bibitem{Haykin} Haykin, Simon S. ``Neural Networks: A Comprehensive Foundation,'' Prentice Hall. (1999)


\bibitem{Hopfield} J. J. Hopfield, "Neural networks and physical systems with emergent collective computational abilities", PNAS 79(8) pp. 2554-2558, 1982

\bibitem{Hopfield2} J. J. Hopfield, "Neurons with graded response have collective computational properties like those of two-state neurons", PNAS, 81(10), pp. 3088-3092, 1984

  \bibitem{Carleo:2019ptp}
G.~Carleo, I.~Cirac, K.~Cranmer, L.~Daudet, M.~Schuld, N.~Tishby, L.~Vogt-Maranto and L.~Zdeborova, ``Machine learning and the physical sciences,''
Rev. Mod. Phys. \textbf{91}, no.4, 045002 (2019)



\bibitem{Jaynes}  E.~ T.~Jaynes, ``Information Theory and Statistical Mechanics,'' Physical Review. Series II. 106 (4): 620-630, (1957)

\bibitem{Jaynes2}  E.~T.~Jaynes, "Information Theory and Statistical Mechanics II," Physical Review. Series II. 108 (2): 171-190, (1957)


\bibitem{Prigogine} Prigogine, I. ``Etude Thermodynamique des ph\'enom\'enes irr\'eversibles''. Desoer, Li\'ege,  (1947)

\bibitem{Klein} M. ~J. ~Klein, P.~ H.~ E. ~ Meijer, ``Principle of minimum entropy production.'' Phys. Rev. 96: 250-255, (1954)


\bibitem{entropic}
  V.~Vanchurin,
  ``Entropic Mechanics: towards a stochastic description of quantum mechanics,''
  Found.\ Phys.\  {\bf 50}, no. 1, 40 (2019)
  
  
\bibitem{emergentgravity}
    V.~Vanchurin,
  ``Covariant Information Theory and Emergent Gravity,''
  Int.\ J.\ Mod.\ Phys.\ A {\bf 33}, no. 34, 1845019 (2018)
  
  
  
\bibitem{Cybenko} Cybenko, G. (1989) "Approximations by superpositions of sigmoidal functions", Mathematics of Control, Signals, and Systems, 2(4), 303?314  
  
\bibitem{Hornik} Hornik, Kurt. "Approximation capabilities of multilayer feedforward networks." Neural networks 4.2 (1991): 251-257.
  
\bibitem{TensorFlow}  
  M. Abadi, P. Barham, J. Chen, Z. Chen, A. Davis, J. Dean, M. Devin, S. Ghemawat, G. Irving, M. Isard et al., ``Tensorflow: A system for large- scale machine learning,'' in 12th USENIX Symposium on Operating Systems Design and Implementation (OSDI 16), 2016, pp. 265-283.


\bibitem{MNIST}
Y. LeCun, L. Bottou, Y. Bengio, and P. Haffner, ``Gradient-based learning applied to document
recognition,'' Proceedings of the IEEE, 86(11):2278-2324, 1998.

  \bibitem{Complexity} 
  A.~R.~Brown, D.~A.~Roberts, L.~Susskind, B.~Swingle and Y.~Zhao,
  ``Holographic Complexity Equals Bulk Action?,''
  Phys.\ Rev.\ Lett.\  {\bf 116}, no. 19, 191301 (2016)
  \bibitem{Complexity1} 
  A.~R.~Brown, D.~A.~Roberts, L.~Susskind, B.~Swingle and Y.~Zhao,
  ``Complexity, action, and black holes,''
  Phys.\ Rev.\ D {\bf 93}, no. 8, 086006 (2016)
\bibitem{Complexity2} 
  A.~R.~Brown and L.~Susskind,
 ``Second law of quantum complexity,''
  Phys.\ Rev.\ D {\bf 97}, no. 8, 086015 (2018)

\bibitem{world}
V.~Vanchurin, ``The world as a neural network,'' Entropy 22 (2020) 1210.

\bibitem{quantumness}
M.~I.~Katsnelson and V.~Vanchurin,
``Emergent Quantumness in Neural Networks,''
[arXiv:2012.05082 [quant-ph]].
  
  
 \bibitem{PDG}  
P.~A.~Zyla \textit{et al.} [Particle Data Group],
``Review of Particle Physics,''
PTEP \textbf{2020}, no.8, 083C01 (2020)

 \bibitem{Planck}  
N.~Aghanim \textit{et al.} [Planck],
``Planck 2018 results. I. Overview and the cosmological legacy of Planck,''
Astron. Astrophys. \textbf{641}, A1 (2020)



\bibitem{Maldacena}
  J.~M.~Maldacena,
  ``The Large N limit of superconformal field theories and supergravity,''
  Int.\ J.\ Theor.\ Phys.\  {\bf 38}, 1113 (1999)

\bibitem{Witten} 
  E.~Witten,
  ``Anti-de Sitter space and holography,''
  Adv.\ Theor.\ Math.\ Phys.\  {\bf 2}, 253 (1998)
  
 \bibitem{Susskind}
   L.~Susskind,
  ``The World as a hologram,''
  J.\ Math.\ Phys.\  {\bf 36}, 6377 (1995)
  
  
  
  
\bibitem{Jacobson} 
  T.~Jacobson,
  ``Thermodynamics of space-time: The Einstein equation of state,''
  Phys.\ Rev.\ Lett.\  {\bf 75}, 1260 (1995)
  
  
  \bibitem{Verlinde} 
  E.~P.~Verlinde,
  ``On the Origin of Gravity and the Laws of Newton,''
  JHEP {\bf 1104}, 029 (2011)



\bibitem{Ryu}
  S.~Ryu and T.~Takayanagi,
  ``Holographic derivation of entanglement entropy from AdS/CFT,''
  Phys.\ Rev.\ Lett.\  {\bf 96}, 181602 (2006)
  
 
\bibitem{Swingle}
B.~Swingle,
  ``Entanglement Renormalization and Holography,''
  Phys.\ Rev.\ D {\bf 86}, 065007 (2012)

  \bibitem{Almheiri} 
  A.~Almheiri, X.~Dong and D.~Harlow,
 ``Bulk Locality and Quantum Error Correction in AdS/CFT,''
  JHEP {\bf 1504}, 163 (2015)
  
  
  

\bibitem{Everett}  H.~Everett, ``Relative State Formulation of Quantum Mechanics,'' Reviews of Modern Physics. 29 (3): 454-462, (1957)

\bibitem{Bohm} D.~Bohm, ``A Suggested Interpretation of the Quantum Theory in Terms of 'Hidden Variables' I,'' Physical Review. 85 (2): 166-179, (1952)


\bibitem{Bohr} Bohr, N. (1928), ``The Quantum Postulate and the Recent Development of Atomic Theory,'' Nature. 121 (3050): 580?590
  
\bibitem{Bell} J.~ Bell, ``On the Einstein Podolsky Rosen Paradox,'' Physics. 1 (3): 195-200, (1964)

\bibitem{Measure} V.~Vanchurin, A.~Vilenkin and S.~Winitzki,
 ``Predictability crisis in inflationary cosmology and its resolution,''
  Phys.\ Rev.\ D {\bf 61}, 083507 (2000)


\bibitem{Tononi}
 G.~Tononi, ``Consciousness as integrated information: a provisional manifesto,'' Biol Bull 215: 216-242,  (2008)

\bibitem{Dvali} 
  G.~Dvali,
 ``Black Holes as Brains: Neural Networks with Area Law Entropy,''
  Fortsch.\ Phys.\  {\bf 66}, no. 4, 1800007 (2018)

\bibitem{graphflow} 
  V.~Vanchurin,
  ``Information Graph Flow: a geometric approximation of quantum and statistical systems,''
  Found.\ Phys.\  {\bf 48}, no. 6, 636 (2018)

\bibitem{spacetime}
  V.~Vanchurin,  ``A quantum-classical duality and emergent space-time,''
10th Mathematical Physics Meeting 2019, 347-366
  
  
\bibitem{Onsager}
Onsager, L.
``Reciprocal relations in irreversible processes, I''.
  Physical Review. 37 (4) 405-426 (1931)

  
\end{thebibliography}
\end{document}